\documentclass[letterpaper, journal]{IEEEtran}
\IEEEoverridecommandlockouts
\usepackage{cite}
\usepackage{amsmath,amssymb,amsfonts}
\usepackage{algorithmic}
\usepackage{graphicx}
\usepackage{textcomp}
\usepackage{xcolor}
\usepackage{booktabs}
\usepackage{multirow}
\usepackage{microtype}
\usepackage{stfloats}
\usepackage{threeparttable} 
\usepackage{comment}
\usepackage[hidelinks]{hyperref}
\usepackage{adjustbox}

\usepackage{soul}
\sethlcolor{yellow}

\soulregister\cite7
\soulregister\ref7
\soulregister\eqref7

\newcommand{\rev}[1]{#1}

\newcommand{\revnew}[1]{#1}

\title{Physics-Embedded Neural ODEs for Learning
Antagonistic Pneumatic Artificial Muscle Dynamics}

\author{Xinyao Wang and Jonathan Realmuto%
  \thanks{This work was supported by NSF under award CMMI-2221315. The authors
    are with the Dept. of Mechanical Engineering, University of California,
    Riverside, CA, USA. Corresponding author J.R., {\tt\small
    jrealmut@ucr.edu}. Code available at: \url{https://github.com/jonreal/pneu-sim/tree/v0.1.0}.
    This work has been submitted to the IEEE for possible publication.
    Copyright may be transferred without notice, after which this version may no
    longer be accessible.}}

\begin{document}

\maketitle

\begin{abstract}

Pneumatic artificial muscles (PAMs) enable compliant actuation for soft
wearable, assistive, and interactive robots. \revnew{When arranged
antagonistically, PAMs can vary joint stiffness through co-contraction, but
their coupled, nonlinear, and rate-dependent dynamics are difficult to model.}
This paper presents a hybrid neural ordinary differential equation (Neural ODE)
framework that embeds physical structure into a learned model of antagonistic
PAM dynamics. The formulation combines parametric joint mechanics and pneumatic
state dynamics with a neural network force component that captures antagonistic
coupling and rate-dependent hysteresis. \rev{The forward model was trained on
29 selected co-contraction conditions and predicted joint motion and chamber
pressures over 196 held-out conditions with a \revnew{median $R^2$ of 0.89}.}
An inverse formulation, derived from the learned dynamics, computes
\revnew{feedforward pressure commands offline for desired motion, stiffness,
and payload profiles. With the valves closed, the mean identified stiffness was
within 3\% of the commanded value from 136 to 176~N/mm and remained consistent
across operating velocities, unlike a static model. In a payload-holding task,
the model predicted the displacement under payloads of up to 173~N within 14\%,
and its feedforward plans removed 93--95\% of it, against 81--82\% for the
static model.}

\end{abstract}


\section{Introduction}
\label{sec:intro}

\noindent Soft pneumatic actuators are widely used in wearable and assistive
robots due to their inherent compliance, lightweight construction, and ability
to generate human-compatible forces~\cite{asbeck2015biologically,
realmuto2019robotic, sridar2017development,bardi2022upper}. These properties
enable safe physical interaction in rehabilitation, mobility assistance, and
human augmentation applications. However, soft pneumatic actuators are
nonlinear, hysteretic, and strongly pressure-dependent; their behavior varies
with actuator design and operating conditions~\cite{zhang2021comprehensive,
xavier2022soft}. Accurately capturing these dynamics remains a fundamental
challenge for model-based control.

Among soft pneumatic actuators, the McKibben-type pneumatic artificial muscles
(PAMs) are widely studied due to their contractile behavior, high
force-to-weight ratio, and well-characterized braid-bladder
geometry~\cite{schulte1961characteristics, chou1996measurement}. In robotic
applications, PAMs are often arranged antagonistically to enable variable joint
stiffness through co-contraction, analogous to human musculoskeletal
control~\cite{hogan1984adaptive,feldman1986once}. This capability supports
direct impedance modulation by antagonist co-activation~\cite{ariga2012novel,
todtheide2015antagonistic}, rather than rendering impedance purely through
high-bandwidth feedback~\cite{hogan1985impedance}. However, the interaction
between opposing muscles through a shared joint transmission introduces coupled
pressure, force, and motion dynamics that further complicates model-based
control of PAM joints.

\rev{Classical analytical PAM models capture quasi-static force-length-pressure
behavior \cite{chou1996measurement}, while empirical and semi-empirical dynamic
models improve force prediction in specific operating regimes
\cite{doumit2009analytical, hovsovsky2012dynamic, kang2009dynamic}. For
antagonistic configurations, prior work includes quasi-static hysteresis and
creep models \cite{minh2012modeling} and low-order dynamic joint models
\cite{hovsovsky2015enhanced}. Learning-based approaches include neural networks
for \revnew{soft pneumatic actuator} prediction \cite{gillespie2018learning,
hyatt2019model}, GRU models \cite{chung2014empirical}, Koopman/lifted-linear
representations \cite{korda2018linear}, and hybrid physics-informed learning
for soft pneumatic systems \cite{sun2022physics, wang2025learning,
wang2025dynamic, magdy2025hybrid, wang2026data}. \revnew{Recent models combine
physics and learning for soft robots and PAMs
\cite{huang2025highprecision,wang2024online,wang2025physics,yan2025robust},
including a physics-informed Neural ODE \cite{youm2026physics}. A latent-state
Neural ODE was also learned from input--output data of an antagonistic PAM
joint at fixed mean pressure for position control \cite{bachhuber2024soft}.
These models target motion or force tracking.}}

\revnew{Model predictive~\cite{zhou2025model} and sliding
mode~\cite{zhao2020active} control have been used for motion tracking in
antagonistic PAM joints. Simultaneous position and stiffness control has been
achieved through decoupling~\cite{trumic2021decoupled}, PI feedback using a
static force model~\cite{takeda2025encrypted}, and Gaussian processes mapping
position and stiffness targets to actuator pressures~\cite{habich2023learning}.
The latter two approaches rely on static models and do not validate stiffness
during motion.} For stiffness estimation, unscented Kalman filters provide
sensorless angle and torque estimation in antagonistic PAM joints
\cite{shin2021detailed,shin2022sensorless}.

\revnew{To our knowledge, no dynamic model has predicted and prescribed joint
stiffness from motion and pressure responses to known loads, without
muscle-force or stiffness targets, and validated stiffness during motion.
Table~\ref{tab:si_prior} compares the closest prior work. We address this gap
with a hybrid Neural ODE~\cite{chen2018neural} combining explicit joint and gas
dynamics with a neural network mapping joint position, velocity, and chamber
air masses to net antagonistic force. The equivalent reflected mass, gas and deformation
coefficients, and network weights are jointly learned by minimizing
state-trajectory errors, with measured motor force as an input and neither
muscle force nor stiffness as a target. Physics is embedded in the governing
equations rather than imposed through loss penalties. Joint stiffness follows
from the position derivative of the learned force, enabling simultaneous motion
and stiffness prescription without separate stiffness identification.}

Our objective is a continuous-time dynamic model of an antagonistic PAM joint
with sufficient physical fidelity to support stiffness-aware feedforward
planning. Given a desired joint trajectory $x_d(t)$ and time-varying stiffness
profile $K_d(t)$, antagonistic muscle air masses $(m_f(t), m_e(t))$ are
synthesized offline from the learned dynamics to reproduce the target motion
and stiffness. 

\rev{In wearable and assistive robots, stiffness modulation can reduce
resistance during user-guided motion \revnew{and increase support against
loads. Here, the model is verified through the joint response to external loads
during tracking and in a payload-holding task.}}

The main contributions are: \rev{(1) a hybrid Neural ODE that embeds
\revnew{joint mechanics} and pressure dynamics for antagonistic PAM modeling;
\revnew{(2) a learned scalar force model that represents antagonistic coupling
and rate-dependent hysteresis through its dependence on position and velocity,
without a hysteresis state;}} and \revnew{(3) an inverse synthesis procedure
that converts the model into offline feedforward pressure commands for
prescribed motion and stiffness, used here to verify the model on an
antagonistic PAM joint, including a payload-holding task.} \rev{The formulation
is expressed in the \revnew{tendon coordinate}, so the learned force model is
not tied to the rotary implementation.}

\section{Hybrid Neural ODE Model}
\label{sec:hnode}

This section develops the hybrid Neural ODE model for the antagonistic PAM
joint (Fig.~\ref{fig:system_model}) under closed-valve conditions, where
chamber air masses remain constant. Valve and mass-flow dynamics are outside
the model so that the \rev{learned scalar force model} represents
mechanical-state-dependent coupling, \revnew{rate-dependent hysteresis}, and
stiffness variation independent of how the chamber loading is established.

\subsection{System States and Inputs}

The antagonistic PAM joint is modeled using the \revnew{tendon coordinate $x$
(the tendon displacement at the pulley),} obtained from the joint rotation $q$
through
\[
x=r_pq,
\]
where $r_p$ is the pulley radius
(Fig.~\ref{fig:system_model}{\textbf{a}},{\textbf{b}}). This coordinate is used
because the PAM forces act directly \revnew{along the tendons}. The system
state vector is defined as
\[
  \mathbf{x}(t) =
\begin{bmatrix}
  x(t) & \dot{x}(t) & P_f(t) & P_e(t)
\end{bmatrix}^{\!\top},
\]
where $x$ and $\dot{x}$ are the joint displacement and velocity, and $P_f$ and
$P_e$ are the absolute flexor and extensor chamber pressures.

During closed-valve interaction, when each PAM chamber is sealed, the internal
masses are constant; therefore, we treat the antagonistic chamber mass pair as
an exogenous input to the model:
\[
\mathbf{u}(t) =
\begin{bmatrix}
m_f(t) & m_e(t)
\end{bmatrix}^{\!\top}.
\]

\subsection{Joint Dynamics}
\label{subsec:joint_dyn}

\begin{figure}[t]
    \centering
    \includegraphics[width=\linewidth,trim=0 0.0 0 5.8,clip]{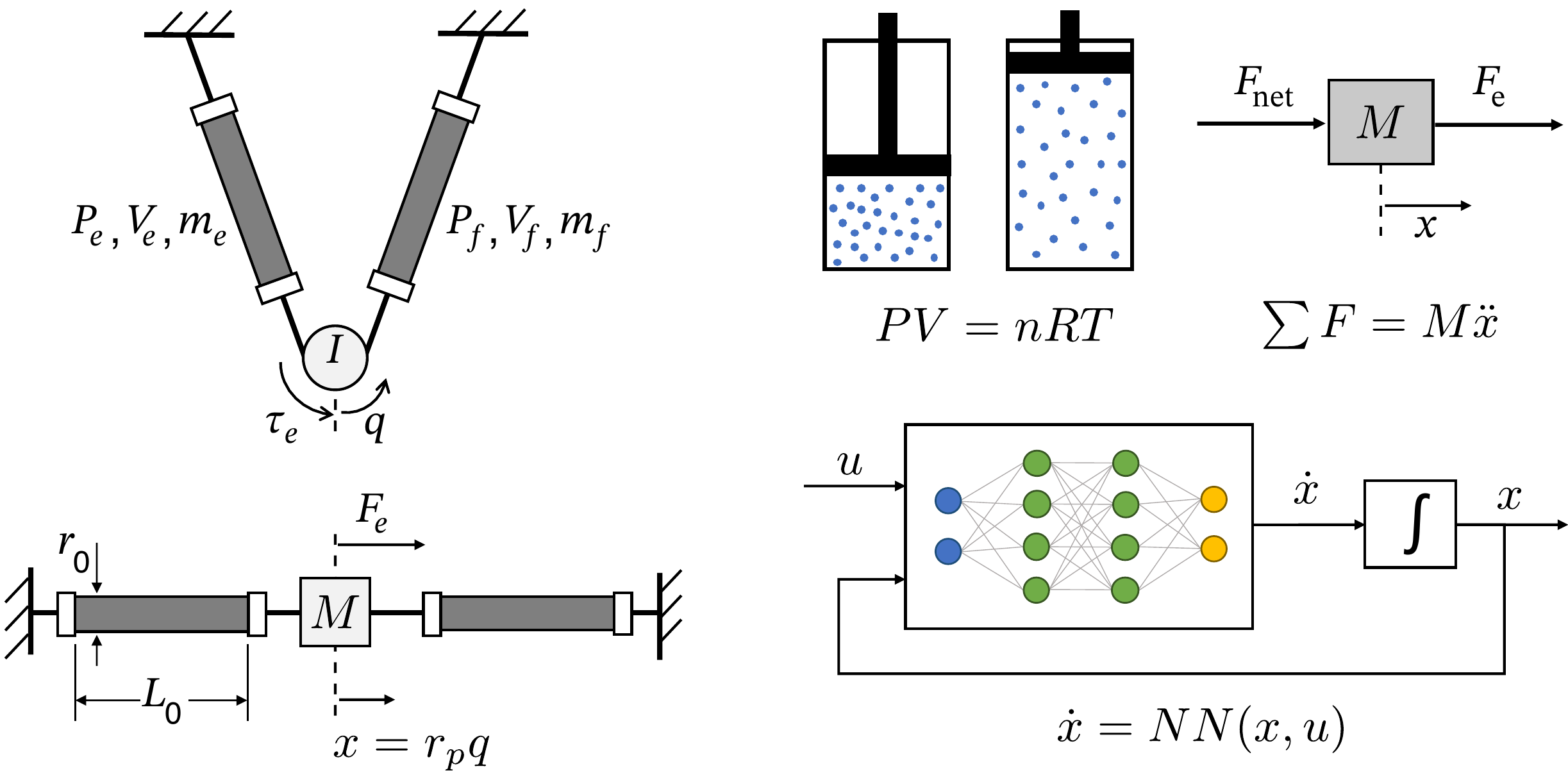}
    \put (-240,115){\textbf{a}}
    \put (-250,40){\textbf{b}}
    \put (-130,115){\textbf{c}}
    \put (-120,60){\textbf{d}}
\caption{%
\rev{System schematic and modeling ingredients. {\textbf{a}} Antagonistic PAM
joint with flexor (f) and extensor (e) chamber pressures, volumes, and air
masses, pulley inertia $I$, joint angle $q$, and external torque $\tau_e$.
{\textbf{b}} \revnew{Equivalent model in the tendon coordinate $x=r_pq$:
equivalent reflected mass $M$,} external force $F_\mathrm{e}$, and nominal
muscle geometry $(r_0,L_0)$. {\textbf{c}} Embedded physics: ideal-gas law and
Newtonian mechanics. {\textbf{d}} Neural ODE concept: a neural network (NN)
maps the state and an exogenous input $u$ to the state derivative, which is
integrated to predict the state trajectory. In our hybrid model, the network
represents only the antagonistic force term within the physics-based dynamics
of {\textbf{a}}--{\textbf{c}}.}}\label{fig:system_model}
\end{figure}

The antagonistic joint is modeled as a single-degree-of-freedom \revnew{system
in the tendon coordinate} (Fig.~\ref{fig:system_model}{\textbf{c}}). The joint
dynamics are expressed as
\begin{equation} \label{eq:joint_dyn}
% V3 R4-C3, symbol m -> M (old): m \ddot{x} = F_{\mathrm{e}} + F_{\mathrm{net}},
M \ddot{x} = F_{\mathrm{e}} + F_{\mathrm{net}},
\end{equation}
\rev{\revnew{where $M=I/r_p^2$ is the equivalent reflected mass, the joint
inertia $I$ expressed in the tendon coordinate through the pulley radius
$r_p$,} $F_e$ represents the externally applied force, and $F_{\mathrm{net}}$
is the net interaction force generated by the antagonistic PAM pair.}

\rev{Rather than decomposing $F_{\mathrm{net}}$ into separate analytical PAM
force models, this work represents it using a learned scalar force model
introduced in Section~\ref{subsec:hybrid_model}.} This formulation allows the
model to capture unmodeled effects such as rate-dependent hysteresis and
antagonistic coupling without relying on explicit force-length-pressure
mappings or actuator-specific stiffness parameters.

\subsection{PAM Pressure Dynamics}
\label{subsec:pressure_dynamics}

For actuator $i\in\{f,e\}$, let $P_i$ denote the absolute chamber pressure,
$m_i$ the chamber air mass, and $V_i(x)$ the configuration-dependent chamber
volume. Under an isothermal ideal-gas approximation
(Fig.~\ref{fig:system_model}{\textbf{c}}),
\begin{equation}
P_i(t) = C\,\frac{m_i}{V_i(x(t))},
\label{eq:mass_pressure_relation}
\end{equation}
where $C=RT/m_{\mathrm{air}}$, $R$ is the universal gas constant, $T$ is the
absolute temperature, and $m_{\mathrm{air}}$ is the molar mass of air. Thus,
the same chamber air mass produces different pressures as the chamber volume
changes with joint displacement. This motivates using chamber mass rather than
instantaneous pressure as an input to the learned force model.

Differentiating \eqref{eq:mass_pressure_relation} gives
\begin{equation}
\dot{P}_i
=
C\left(
\frac{\dot{m}_i}{V_i(x)}
-
\frac{m_i\,\dot{V}_i(x,\dot{x})}{V_i(x)^2}
\right),
\label{eq:pressure_dynamics}
\end{equation}
where $\dot{m}_i$ is the mass-flow rate into the chamber and
$\dot{V}_i(x,\dot{x})$ is the chamber-volume rate. Under the closed-valve
conditions considered in this work, $\dot{m}_i=0$, while pressure continues to
vary as joint motion changes the chamber volume
(Fig.~\ref{fig:system_model}{\textbf{c}}). Temperature variation, leakage, and
other unmodeled pneumatic effects are not measured explicitly; therefore, $C$
is treated as a learnable parameter during identification.

\subsection{Volume-Displacement Relationship}
\label{subsec:volume}

The inextensible braid couples PAM axial length and radial expansion. Modeling
each PAM as a cylinder with displacement-dependent radius gives the chamber
volumes
\begin{equation}
V_f(x) = \pi r_f(x)^2\,(L_0 - x), \quad
V_e(x) = \pi r_e(x)^2\,(L_0 + x),
\label{eq:volume_model}
\end{equation}
where $L_0$ is the nominal (rest) muscle length, and $r_f(x)$ and $r_e(x)$
denote the effective radii of the flexor and extensor PAMs.

The radius-length coupling follows from the inextensible braid constraint.
Assuming a linear coupling about the nominal configuration gives
\[
r_f(x) = r_0 + \nu\frac{r_0}{L_0}x, \quad
r_e(x) = r_0 - \nu\frac{r_0}{L_0}x,
\]
where $r_0$ is the nominal PAM radius and $\nu$ is a deformation coefficient
determined by braid geometry. \revnew{This volume relation is the only part of
the model specific to the braided muscle; another pneumatic actuator would
require its own $V_i(x)$.} Differentiating \eqref{eq:volume_model} gives the
time derivatives of volume:
\[
\begin{aligned}
\dot{V}_f(x,\dot{x})
&= \pi \dot{x}\Bigl[ 2r_f(x)(L_0 - x)\nu \frac{r_0}{L_0} - r_f(x)^2 \Bigr], \\
\dot{V}_e(x,\dot{x})
&= -\pi \dot{x}\Bigl[ 2r_e(x)(L_0 + x)\nu \frac{r_0}{L_0} - r_e(x)^2 \Bigr].
\end{aligned}
\label{eq:volume_rate}
\]

\subsection{Forward Hybrid Neural ODE Model}
\label{subsec:hybrid_model}

\rev{The antagonistic PAM dynamics are constructed by combining the parametric
joint mechanics in~\eqref{eq:joint_dyn} with the pressure dynamics
in~\eqref{eq:pressure_dynamics}. The resulting Neural ODE vector field is
integrated to predict the system-state trajectory
(Fig.~\ref{fig:system_model}{\textbf{d}}):}
\begin{equation}
  \dot{\mathbf{x}} =
\begin{bmatrix}
\dot{x} \\
\ddot{x} \\
\dot{P}_f \\
\dot{P}_e
\end{bmatrix}
=
\begin{bmatrix}
\dot{x} \\
\dfrac{1}{M}\!\left(F_\mathrm{e} + f_\theta(x,\dot{x},m_f,m_e)\right) \\
C\!\left(\dfrac{\dot{m}_f}{V_f(x)} - \dfrac{m_f \dot{V}_f(x,\dot{x})}{V_f(x)^2}\right) \\
C\!\left(\dfrac{\dot{m}_e}{V_e(x)} - \dfrac{m_e \dot{V}_e(x,\dot{x})}{V_e(x)^2}\right)
\end{bmatrix},
\label{eq:state_dynamics}
\end{equation}
\rev{Here, $f_\theta$ is the learned scalar force model that approximates the
unknown net interaction force $F_{\mathrm{net}}$ in~\eqref{eq:joint_dyn}, with
$\theta$ denoting the neural network parameters.} \revnew{The force $f_\theta$
is memoryless, but its dependence on $\dot{x}$ captures rate-dependent
hysteresis: different loading and unloading forces at the same displacement.
Without a hysteresis state, loading-history effects are not represented.}

For compactness, \eqref{eq:state_dynamics} is written as
\begin{equation}
  \dot{\mathbf{x}} = \mathbf{f}_{\Theta}(\mathbf{x},\mathbf{u},F_\mathrm{e}),
    \label{eq:hybrid_ode_compact}
\end{equation}
\rev{where $\mathbf{f}_{\Theta}$ denotes the complete Neural ODE vector field and the learnable parameter set is}
\[
\Theta = \{\,M,\; C,\; \nu,\; \theta\,\}.
\]
\rev{Here, \revnew{$M$}, $C$, $\nu$, and $\theta$ denote the \revnew{equivalent
reflected mass}, gas coefficient, deformation coefficient, and neural network
weights. The Neural ODE formulation treats the physics-embedded model as a
continuous-time trainable vector field, allowing trajectories to be integrated
on the measured time grid rather than using a fixed-step discrete predictor.
Because the learned scalar force model is separated from the rollout step,
stiffness can be computed directly from $\partial f_\theta/\partial x$ at the
desired operating point instead of from a discrete-time state-transition map.}

\subsection{Feedforward Input Synthesis via Constrained Optimization}
\label{subsec:inverse_model}

\begin{figure}[t]
    \centering
    \includegraphics[width=\columnwidth,trim=0 3.8 0 3.8,clip]{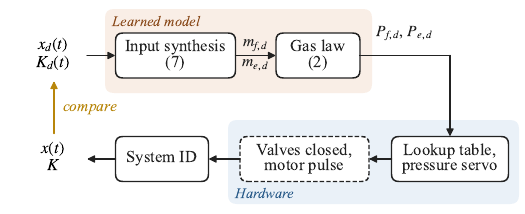}
    \caption{%
\revnew{Verification pipeline. The learned model synthesizes air masses for
target motion $x_d(t)$ and stiffness $K_d(t)$ \eqref{eq:synthesis}, then
converts them to pressure commands $P_{f,d}$ and $P_{e,d}$ using the gas law
\eqref{eq:mass_pressure_relation}. During testing, a pressure servo tracks
these commands from a lookup table. In brief closed-valve windows (dashed), the
servo pauses while a motor pulse probes stiffness at constant air mass.
Identified stiffness $K$ and measured $x(t)$ are compared with the
targets.}}\label{fig:offline2online}
\end{figure}

\revnew{To verify the learned dynamics, model-synthesized pressure commands for
prescribed motion and stiffness are tracked on the hardware and compared with
measured responses (Fig.~\ref{fig:offline2online}).} Given a desired joint
trajectory $x_d(t)$ with corresponding velocity $\dot{x}_d(t)$, acceleration
$\ddot{x}_d(t)$, and a target stiffness profile $K_d(t)$, we synthesize a
feedforward antagonistic input at each time step by solving a constrained
optimization problem for the chamber air masses $(m_f,m_e)$. The optimization
is evaluated at the desired operating point $(x_d(t),\dot{x}_d(t))$ and
enforces dynamic consistency with $\ddot{x}_d(t)$.

The stiffness $K(t)$ is defined as a \emph{translational stiffness} \revnew{in
the tendon coordinate} (N/mm),
\[
K = \frac{\partial f_\theta} {\partial x}.
\]
Because the \rev{learned scalar force model} $f_\theta(\cdot)$ uses
piecewise-linear activations (LeakyReLU), the derivative $\partial f_\theta /
\partial x$ is piecewise constant and discontinuous at activation boundaries.
\rev{To obtain a smooth stiffness estimate at a given operating point, we
approximate the local slope using a smoothed second-order central-difference
estimate:}
\[
\label{eq:stiffness_fd}
\begin{aligned}
  \hat{K}(x, &\dot{x}, m_f, m_e) =\;
\big( \,f_\theta(x+2h, \dot{x}, m_f, m_e) \\
& + f_\theta(x+h, \dot{x}, m_f, m_e)
- f_\theta(x-h, \dot{x}, m_f, m_e) \\ &
- f_\theta(x-2h, \dot{x}, m_f, m_e)\,\big)
 / (6h)
\end{aligned}
\]
where \revnew{$h=1$~mm} is the increment used to approximate the local slope of
the force-displacement relationship at the operating point. \rev{This averages
central-difference slopes at $h$ and $2h$, reducing sensitivity to local
LeakyReLU slope discontinuities while retaining second-order accuracy in smooth
regions.}

At each time sample $t_k$, given the desired trajectory point $(x_{d,k},
\dot{x}_{d,k}, \ddot{x}_{d,k})$ and target stiffness $K_{d,k}$, we solve for
the chamber masses $(m_f, m_e)$ via
\begin{equation} \label{eq:synthesis}
\begin{aligned}
\min_{m_f,m_e}\quad &\big(\hat{K}(x_{d,k}, \dot{x}_{d,k}, m_f, m_e) - K_{d,k}\big)^2 \\
\text{s.t.}\quad &\big|M\ddot{x}_{d,k} - f_\theta(x_{d,k}, \dot{x}_{d,k}, m_f,
m_e)\big| \le \varepsilon,
\end{aligned}
\end{equation}
where the constraint enforces dynamic consistency with the desired
acceleration. \revnew{The solver did not always meet the tight tolerance
$\varepsilon=0.001$~mN, but residuals remained below 0.2~mN. Increasing
$\varepsilon$ from $10^{-3}$ to $10^{2}$~mN changed simulated tracking error by
less than 0.3~$\mu$m while reducing synthesis time up to tenfold
(Fig.~\ref{fig:si_eps}). Thus, a looser tolerance suffices.}

This procedure yields an optimal antagonistic mass pair, \(\mathbf{u}^{*}(t) =
[\,m_f^{*}(t)\;\; m_e^{*}(t)\,]^\top\), that is dynamically consistent with the
desired motion while matching the target local stiffness at the operating
point. The key intuition is that, under sealed interaction, air mass is the
invariant chamber quantity: joint motion changes pressure through volume
change, whereas the mass parameterizes the antagonistic loading state that
determines the force-displacement slope.

\section{Model Identification and Training}
\label{sec:parameter_estimation}

This section describes the learning objective and training procedure used to
identify the parameters of the hybrid Neural ODE introduced
in~\eqref{eq:state_dynamics} in Section~\ref{subsec:hybrid_model}. Physical
parameters and neural network weights are learned jointly from experimental
trajectories by minimizing a trajectory-level prediction error.

\subsection{Training Objective}

The predicted state trajectory $\hat{\mathbf{x}}(t) = [\hat{x},\,
\hat{\dot{x}},\, \hat{P}_f,\, \hat{P}_e]^\top$ is obtained by integrating the
hybrid Neural ODE~\eqref{eq:hybrid_ode_compact} forward in time:
\[
\dot{\hat{\mathbf{x}}} = \mathbf{f}_\Theta(\hat{\mathbf{x}}, \mathbf{u}, F_\mathrm{e}), \quad \hat{\mathbf{x}}(0) = \mathbf{x}_0,
\]
where $\mathbf{x}_0$ is the measured initial state, $\mathbf{u} = [m_f,
m_e]^\top$ are the chamber masses (constant under closed-valve conditions), and
$F_e(t)$ is the applied external force.

\rev{The parameters $\Theta$ are identified by minimizing the error between the
predicted state trajectory $\hat{\mathbf{x}}(t)$ and measured state trajectory
$\mathbf{x}(t)$. Since $\mathbf{x}=[x,\dot{x},P_f,P_e]^\top$, the loss is:}
\[
\begin{aligned}
\mathcal{L}
\;=\;&
100 \cdot \frac{1}{N}\sum_{k=1}^{N}
\Big[
(\hat{x}_k - x_k)^2
+
(\hat{\dot{x}}_k - \dot{x}_k)^2
\Big] \\
&+
\frac{1}{N}\sum_{k=1}^{N}
\Big[
(\hat{P}_{f,k} - P_{f,k})^2
+
(\hat{P}_{e,k} - P_{e,k})^2
\Big],
\end{aligned}
\]
where $N$ denotes the number of time samples in the trajectory. The weighting
coefficients are selected based on the numerical scale of the state variables,
with joint displacement and velocity expressed in millimeters (mm) and
millimeters per second (mm/s), and pressures expressed in kilopascals (kPa).

\subsection{Staged Training Procedure}
\label{subsec:stagetrain}

\rev{Direct training from random initialization was unstable because the
learned force model initially lacked the passive restoring and dissipative
behavior needed for stable ODE rollouts. To stabilize training, a temporary
auxiliary damping term $B\dot{x}$ was added during continuation. The initial
value $B=6500$ kg/s was obtained from prior rotational damping characterization
and converted to the \revnew{tendon coordinate}; $B$ was reduced to zero and
removed from the final model.}

\rev{Each curriculum stage (\revnew{Table~\ref{tab:curriculum}}) was
initialized from the best checkpoint of the previous stage. Training began with
the 170-170~kPa condition (10-10~psi gauge) while $B$ was reduced to zero, then
expanded from symmetric co-contraction pairs to asymmetric and intermediate
pressure pairs to capture stiffness scaling, equilibrium shifts, and
interpolation across the operating space.}

\rev{The same trajectory-level loss was used at all stages. The Neural ODE was
integrated using a fifth-order Tsitouras solver with adjoint sensitivities, and
$f_\theta$ used two 96-neuron hidden layers with LeakyReLU activations. At each
stage, the neural network and learnable physical parameters were optimized
jointly with Adam for up to 5000 epochs \revnew{with initial learning rate
$10^{-2}$, multiplying the learning rate by 0.95 after every 100 epochs without
improvement; the best checkpoint of each stage was retained.} The prescribed
auxiliary damping $B$ was not learned, and the best checkpoint was passed to
the next stage. \revnew{On 8 cores of an AMD EPYC 7713 CPU, one dataset-epoch
costs 15.8--18.5~s, rising with the number of datasets
(Table~\ref{tab:compute}). The $7.0\times10^{4}$ dataset-epochs up to the final
checkpoint correspond to about 333~h on this hardware (at least 108~h on an
Apple M3 Ultra workstation); the stages were launched manually over about two
weeks.}}

\begin{figure}[t]
    \centering

    \begin{minipage}[t]{0.35\linewidth}
        \centering
        \includegraphics[width=\linewidth]{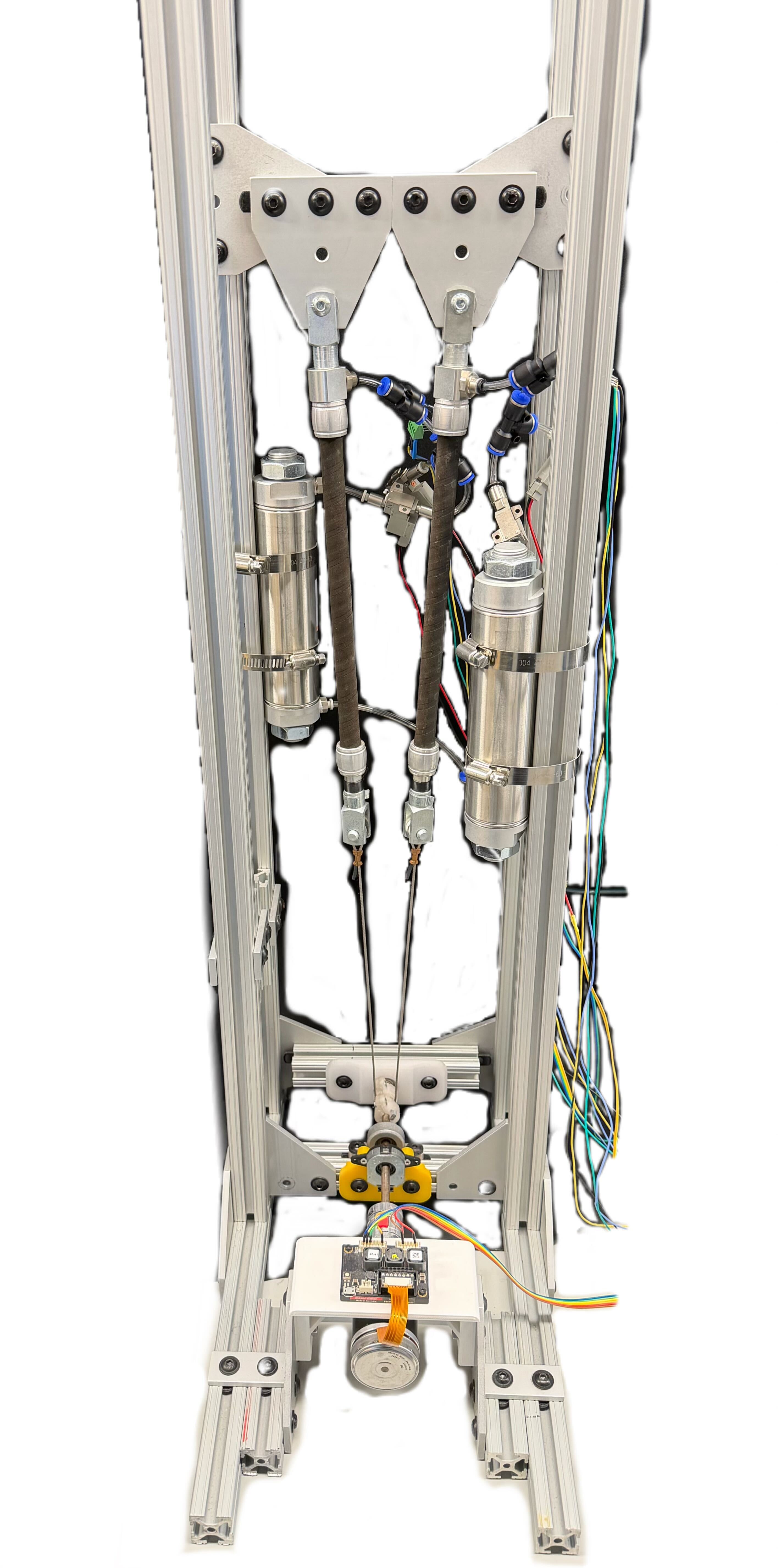}
    \end{minipage}
    \hfill
    \begin{minipage}[t]{0.56\linewidth}
        \centering
        \includegraphics[width=\linewidth]{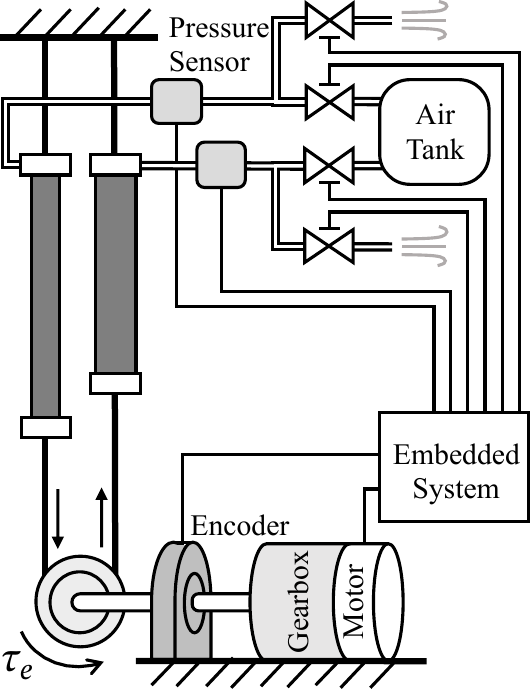}
    \end{minipage}
    \put (-250,170){\textbf{a}}
    \put (-155,170){\textbf{b}}
    \caption{%
    Experimental antagonistic PAM platform.
    {\textbf{a}} Physical setup with antagonistic PAM pair, pulley joint, and DC motor for external torque excitation.
    {\textbf{b}} Pneumatic and sensing schematic: flexor and extensor PAM chambers are independently regulated by three-way solenoid valves, supplied from a common air tank, and measured by inline pressure sensors.}
    \label{fig:setup}
\end{figure}

\subsection{Physical Parameter Initialization}

Initial physical parameters were obtained from datasheets, geometry, and
experimental characterization. The \revnew{equivalent reflected mass $M$} was
computed from the reflected motor/transmission inertia; the deformation
coefficient was initialized from the McKibben braid relation
$\nu=\cot^2(\theta_{braid})$ \cite{chou1996measurement}; and $C$ was
initialized from nominal ideal-gas values. Physical parameters were constrained
positive using a Softplus transformation.

\section{Experimental Setup}
\label{sec:experimentalsetup}

This section describes the experimental platform and data collection procedures
used for model identification and validation of the antagonistic PAM joint.
Quasi-static single-PAM characterization data were used only for offline
air-mass estimation, as described in Section~\ref{subsec:mass_estimation}.
The dynamic model identification and validation experiments were conducted on
the antagonistic joint platform.

\subsection{Experimental Antagonistic Joint Platform}
\label{subsec:platform}

The antagonistic joint platform used for dynamic model identification and
validation is shown in Fig.~\ref{fig:setup}. Two identical PAMs (DMSP-10-200N,
Festo) were mounted in an antagonistic configuration and routed symmetrically
over a pulley to form a single-degree-of-freedom joint. Each PAM was connected
to the joint shaft via a steel tendon secured to a 3D-printed fixture, ensuring
axial loading and symmetric routing.

A brushless DC motor (200142, Maxon Group) with a gearbox (260552, Maxon Group)
was coupled to the joint shaft to apply external torque excitation. \rev{The
motor was not used as the primary actuator of the joint; it was used only to
generate controlled external perturbations for model identification, including
sinusoidal torque excitation for training data collection and short torque
pulses for stiffness identification.} Joint position was measured using an
incremental magnetic encoder (AMT203-V, CUI Devices). Chamber pressures were
measured independently for each PAM using Honeywell 150PG2A3 pressure sensors
mounted near the muscle inlets. A small preload was applied manually to both
PAMs to remove slack and ensure consistent tendon tension.

Table~\ref{tab:system_params} \rev{summarizes measured hardware parameters,
derived quantities, and initialization values.} Parameters marked
${}^{\dagger}$ initialized training and were not fixed identified constants.
\rev{\revnew{The equivalent reflected mass was initialized as
$M^{\dagger}=I/r_p^2$. It is not the mass of a physical body: $I$ is dominated
by the rotor inertia of the motor multiplied by $N^2$, and dividing by $r_p^2$
gives several hundred kilograms. After training, $M=304.65$~kg (a joint inertia
of $1.44\times10^{-2}$~kg$\cdot$m$^2$),} $\nu=3.69$, and $C=8.47\times10^7$
kPa$\cdot$mm$^3$/g, with $C$ unchanged from initialization.}

\begin{figure}[t]
    \centering
    \includegraphics[width=\linewidth]{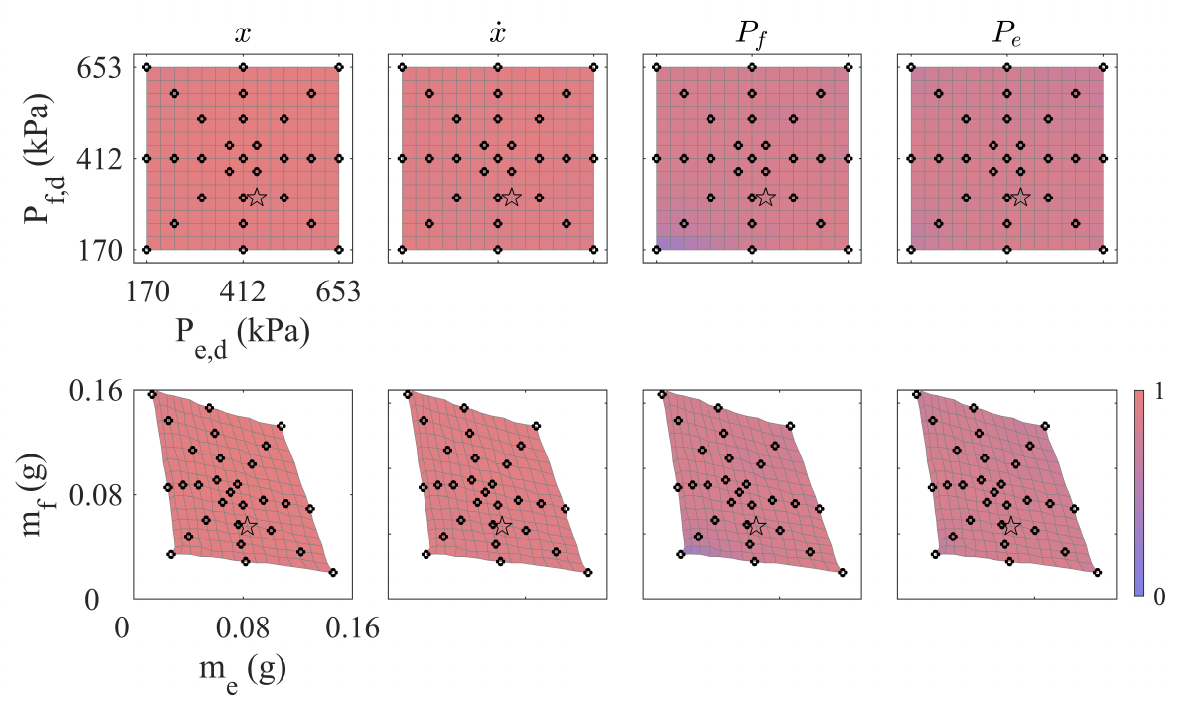}
    \put (-250,140){\textbf{a}}
    \put (-250,70){\textbf{b}}
    \caption{%
    \rev{$R^2$ of the hybrid Neural ODE over 225 co-contraction conditions.
    Circles mark the 29 training datasets; the remaining datasets are held-out validation cases.
    The star marks the condition used for the time-series comparison in Fig.~\ref{fig:response_example}.
    {\textbf{a}} Desired flexor/extensor pressure space $(P_{f,d},P_{e,d})$.
    {\textbf{b}} Corresponding flexor/extensor air-mass space $(m_f,m_e)$.}}
    \label{fig:R2_map}
\end{figure}

\subsection{Pressure Regulation and Embedded System}

Compressed air was supplied from a main storage tank (N310415, PORTER-CABLE)
and passed through an inline pneumatic reservoir (US14227-S0400, SMC) to reduce
pressure fluctuations. The reservoir output was divided into two branches, each
connected to a PAM through a pair of miniature three-way solenoid valves
(SY113-SMO-PM3-F, SMC), enabling independent inflation and exhaust.

Pressure was regulated by discrete valve pulsing. When the pressure error
exceeded a fixed deadband, the controller applied a 10~ms inflation or 15~ms
exhaust pulse, followed by a 20~ms refractory period. These timings were chosen
empirically for reliable valve operation. At the target pressure, both valves
were closed and the PAM chamber was sealed.

A BeagleBone Black embedded controller coordinated valve actuation,
motor torque commands, and synchronized data acquisition for the antagonistic
joint platform. All sensor signals were sampled at 1~kHz and recorded for
offline processing.

\begin{table}[t]
  \centering
  \caption{\rev{System parameters used in the model. Measured quantities come from hardware specifications or direct measurement; derived quantities are computed from measured values; $\dagger$ marks initialization values for training.}}
  \label{tab:system_params}
  \renewcommand{\arraystretch}{1.15}
  \setlength{\tabcolsep}{6pt}
  \begin{tabular}{l l c l}
    \hline
    \textbf{Symbol} & \textbf{Description} & \textbf{Value} & \textbf{Unit} \\
    \hline
    $r_0$ & Nominal PAM radius & $5.00\times 10^{-3}$ & m \\
    $L_0$ & Nominal PAM length & $0.2$ & m \\
    $r_p$ & Pulley radius & $6.875\times 10^{-3}$ & m \\
    $I_{\mathrm{motor}}$ & Motor rotor inertia & $9.25\times 10^{-6}$ & kg$\cdot$m$^{2}$ \\
    $I_{\mathrm{gearbox}}$ & Gearbox inertia & $5.00\times 10^{-7}$ & kg$\cdot$m$^{2}$ \\
    $N$ & Gear ratio & $36{:}1$ & - \\
    $I$ & Equivalent joint inertia & $1.20\times 10^{-2}$ & kg$\cdot$m$^{2}$ \\
% V3 R4-C3; 253.0 is the value the training code used (old):     $m^{\dagger}$ & Equivalent linear mass & $253.9$ & kg \\
    \revnew{$M^{\dagger}$} & \revnew{Equivalent reflected mass} & \revnew{$253.0$} & kg \\
    $\theta_{braid}$ & Braid angle & $25$ & deg \\
    $\nu^{\dagger}$ & Deformation coefficient & $4.60$ & - \\
    $R$ & Gas constant & $8.314$ & J/(mol$\cdot$K) \\
    $T$ & Temperature & $295.15$ & K \\
    $m_{\mathrm{air}}$ & Air molar mass & $2.897\times 10^{-2}$ & kg/mol \\
    $C^{\dagger}$ & Gas coefficient & $8.47\times10^{7}$ & kPa$\cdot$mm$^{3}$/g \\
    \hline
  \end{tabular}
\end{table}

\subsection{Training Excitation}
\label{subsec:input_excitation}

The antagonistic joint was excited using sinusoidal motor torque inputs only.
Three excitation frequencies were used during data collection: 0.5\,Hz, 1\,Hz,
and 2\,Hz. For each frequency, two torque amplitudes were applied by commanding
motor currents of 1\,A (0.918\,N$\cdot$m) and 1.5\,A (1.377\,N$\cdot$m). These
trials were repeated under multiple co-contraction conditions to generate a
dataset covering a range of stiffness configurations for model training and
held-out validation. \revnew{Stiffness predictions were validated in separate
perturbation experiments using 150~ms, 0.5~A torque pulses
(Section~\ref{subsec:stiffness}); these data were not used for training.}

\subsection{Air Mass Estimation}
\label{subsec:mass_estimation}

The chamber air masses $(m_f, m_e)$ are not directly measurable but can be
inferred from pressure and position measurements. Using offline single-PAM
characterization, quasi-static loading and unloading force profiles were
measured across a range of pressures and fitted with polynomial surfaces. These
profiles relate force, displacement, and air mass for each hysteresis branch.

\begin{figure*}[t]
    \centering
    \includegraphics[width=\textwidth,trim=0 6.7 0 6.9,clip]{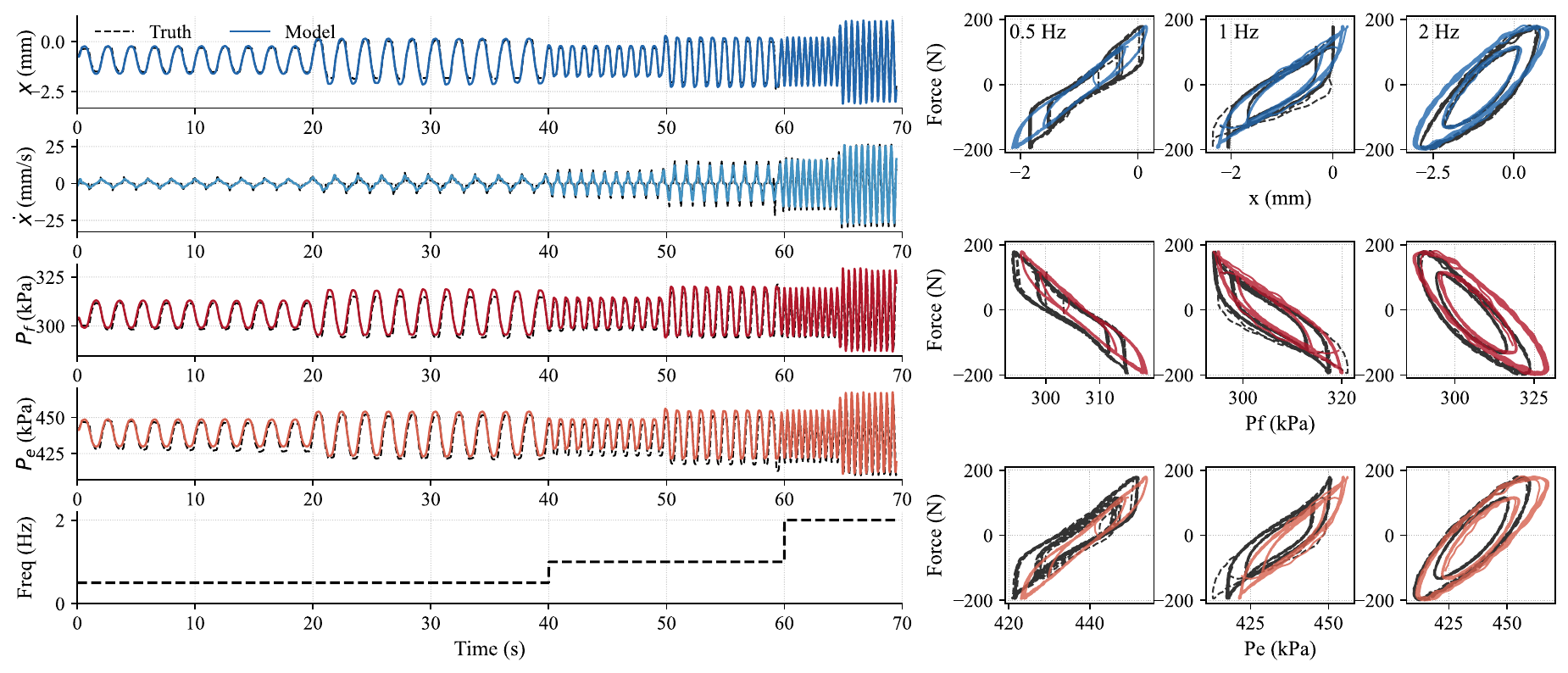}
    \caption{%
Example comparison between the hybrid Neural ODE prediction and experimental
measurements for \revnew{the median held-out condition ($R^2=0.889$) at
$(P_{f,d},P_{e,d})=(308.2,446.1)$~kPa absolute (30, 50~psi gauge).} The model
predicts displacement $x$, velocity $\dot{x}$, and flexor/extensor pressures
$P_f$ and $P_e$. Right panels show measured and predicted hysteresis loops for
$x$-$F$, $P_f$-$F$, and $P_e$-$F$ at 0.5, 1, and 2~Hz.}
\label{fig:response_example}
\end{figure*}

\begin{figure*}[!t]
    \centering
    \includegraphics[width=\textwidth,trim=0 12.3 0 11.0,clip]{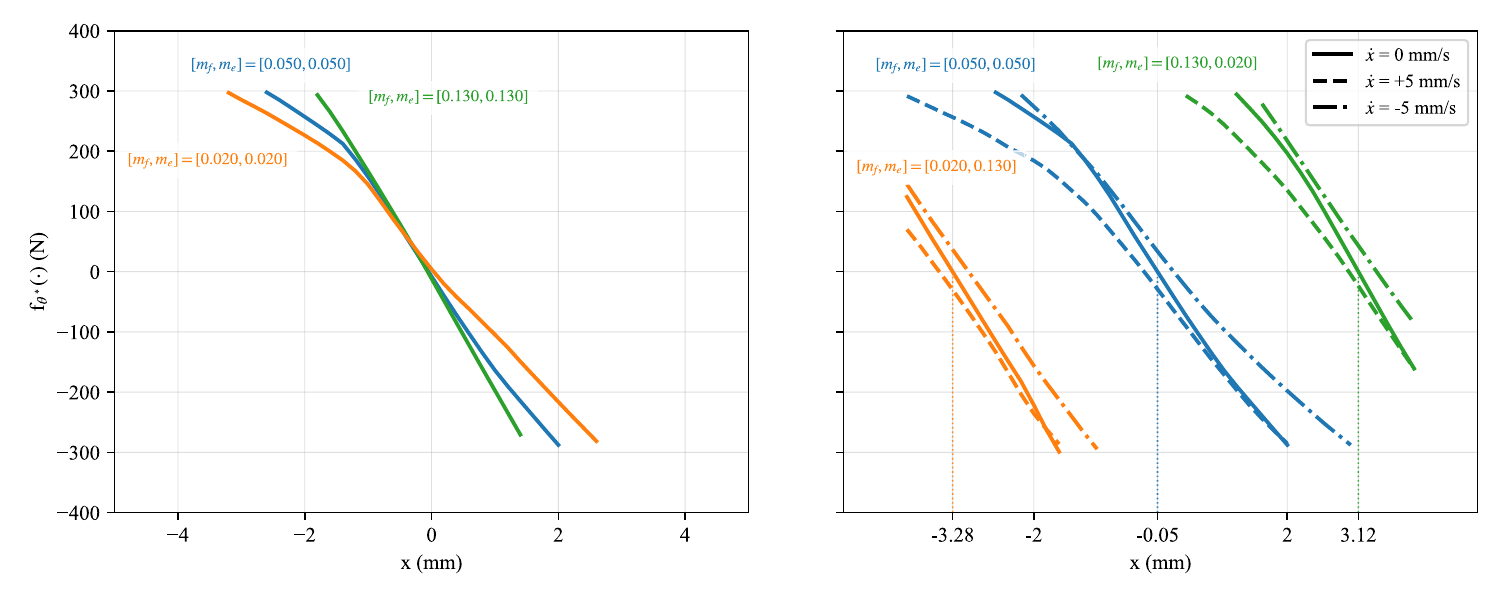}
    \put (-500,190){\textbf{a}}
    \put (-240,190){\textbf{b}}
    \caption{%
\rev{Learned scalar force model} $f_\theta(\cdot)$ versus joint displacement
for discrete flexor/extensor air masses and velocities
$\dot{x}\in\{-5,0,5\}$~mm/s. {\textbf{a}} Symmetric mass pairs $(m_f=m_e)$ show
stiffness changes with total air mass. {\textbf{b}} Asymmetric mass pairs show
equilibrium shifts from flexor-extensor imbalance. Dotted lines mark zero-force
equilibrium positions.} \label{fig:fnet}
\end{figure*}

Because the gas coefficient, $C$, and the deformation coefficient, $\nu$, are
unknown before training, air masses cannot be computed directly from pressure.
Instead, they are estimated at equilibrium configurations where the flexor and
extensor forces balance. Because hysteresis causes the equilibrium position to
differ with velocity direction, masses are estimated at both
loading-to-unloading and unloading-to-loading zero-torque crossings, then
averaged to reduce bias. After training, $C$ and $\nu$ are identified, and air
masses are computed directly via~\eqref{eq:mass_pressure_relation}.

\section{Results}
\label{sec:results_forward}

\rev{This section evaluates forward prediction and learned force behavior
(Section~\ref{subsec:forward_model}), offline inverse motion synthesis
(Section~\ref{subsec:results_inverse}), perturbation-based stiffness synthesis
(Section~\ref{subsec:stiffness}), stiffness consistency relative to an
equilibrium-point model (Section~\ref{subsec:EP_comparison})\revnew{, and a
feedforward load-holding task (Section~\ref{subsec:payload})}. Together, these
experiments assess prediction accuracy, motion synthesis, stiffness modulation,
and the benefit of velocity-dependent force modeling.}

\subsection{Forward Model Accuracy}
\label{subsec:forward_model}

\subsubsection{Proposed Model Validation}

The hybrid Neural ODE forward model was trained using 29 datasets selected from
225 distinct co-contraction conditions. The conditions were defined on a grid
of commanded pressure pairs $(P_{f,\mathrm{d}},P_{e,\mathrm{d}})$
(Fig.~\ref{fig:R2_map}{\textbf{a}}), with the corresponding air masses obtained
using the procedure in Section~\ref{subsec:mass_estimation}
(Fig.~\ref{fig:R2_map}{\textbf{b}}). As detailed in
Section~\ref{subsec:stagetrain}, the selected datasets provide staged coverage
of symmetric and asymmetric conditions across the pressure grid, forming an
``X-shaped'' pattern. The remaining 196 datasets were reserved for evaluation.
\revnew{A single parameter set was applied to all held-out conditions without
retraining.}

Additional datasets along lines with the same total pressure were included to
widen coverage around intermediate co-contraction levels. This intentional,
non-uniform selection maximizes coverage of symmetric and asymmetric regimes
while limiting training size.

For each pressure pair (or equivalently, mass pair), the model predicted the
joint displacement $x$, velocity $\dot{x}$, and PAM pressures $P_f$ and $P_e$.
\revnew{The coefficient of determination $R^2$ was computed for each state over
each trajectory and averaged over the four states.}
\rev{\revnew{Fig.}~\ref{fig:R2_map} summarizes the $R^2$ values across all 225
operating conditions, with the 29 training datasets highlighted by circles.
Forward prediction accuracy was evaluated on the remaining 196 held-out
conditions, with a \revnew{median $R^2$ of 0.89 (mean 0.88, range 0.77--0.91;
median by state, 0.97 for $x$, 0.94 for $\dot{x}$, 0.84 for $P_f$, and 0.82 for
$P_e$).} Lower $R^2$ values mainly occur in low-pressure or highly asymmetric
cases, where the contraction force is small and the motor excitation can cause
one PAM tendon to become slack. When this occurs, the chamber volume changes
less than expected, so the pressure does not decrease consistently with the
closed-valve pressurized-actuator assumption. This failure mode is less
pronounced at higher co-contraction levels, where both PAMs remain tensioned.}
The dataset used for the time-series example in Fig.~\ref{fig:response_example}
is indicated by a star in Fig.~\ref{fig:R2_map}.

Representative time-series results for \revnew{the median held-out condition}
are shown in Fig.~\ref{fig:response_example}. \revnew{The distribution of
held-out $R^2$, with the best and worst conditions, is shown in
Figs.~\ref{fig:si_r2_hist} and \ref{fig:si_best_worst}.} The predicted joint
motion and chamber pressures closely follow the measured trajectories
throughout the excitation cycle. To further examine hysteresis, the
corresponding force versus displacement ($x$-$F$) and pressure versus force
($P_f$-$F$, $P_e$-$F$) loops are also shown for each frequency segment in
Fig.~\ref{fig:response_example}. \revnew{At constant air mass, the $x$-$F$ loop
area measures energy dissipated per cycle, modeled solely through the velocity
dependence of $f_\theta$. Across 196 held-out conditions, median
predicted-to-measured area ratios were 0.81--0.88 at 0.5~Hz, 0.95--0.99 at
1~Hz, and 1.00--1.09 at 2~Hz (ranges across load amplitudes;
Table~\ref{tab:si_loop}). Underprediction at 0.5~Hz is consistent with a
rate-independent loss component not fully captured by the memoryless model.}

\rev{The measured pressure signals show phase delay mainly from the 3 Hz
on-board low-pass filtering. \revnew{For the pressure--force loops in
Fig.~\ref{fig:response_example}, measured pressures were shifted by the
cross-correlation lag between measured displacement and extensor pressure (30,
40, and 41~ms at 0.5, 1, and 2~Hz).} This alignment was not used for training or
quantitative evaluation; the reported $R^2$ values use the original unaligned
trajectories and therefore include this delay.}

\rev{Forward validation was limited to the 0.5, 1, and 2 Hz excitation range
supported by the motor and pressure-regulation hardware. \revnew{Median
held-out displacement $R^2$ increased from 0.959 to 0.978 and 0.982 at 0.5, 1,
and 2~Hz, respectively. At 2~Hz, median four-state $R^2$ fell from 0.905 to
0.820, driven by pressure signals delayed by sensor filtering. Broader
frequency and velocity validation remains future work.}}

\revnew{One further test used an input absent from training
(Fig.~\ref{fig:si_pulse}). For the 60 closed-valve torque pulses of
Section~\ref{subsec:stiffness} (150~ms, about 64~N), the model predicted the
displacement with a median $R^2$ of 0.81; the peak deflection was 12\% low on
average and depended on pulse direction. Both load amplitudes are present in
every training recording; held-out displacement $R^2$ was 0.96 and 0.97 at the
two amplitudes, and the predicted amplitude exceeded the measured one by
9--17\% at 0.5 and 1~Hz.}

\rev{To further examine the learned scalar force model, $f_\theta(\cdot)$ was
evaluated over joint displacement for representative antagonistic air-mass
pairs spanning the operating range and for three velocities $\dot{x} \in
\{-5,\,0,\,5\}~\mathrm{mm/s}$.} The resulting force-displacement curves are
shown in Fig.~\ref{fig:fnet}. Near the neutral configuration, the
force-position relationship is approximately linear, with increasing curvature
at larger displacements and for asymmetric mass distributions. Changing the
relative air masses shifts the equilibrium joint position, while increased
co-contraction increases the local slope, corresponding to higher effective
stiffness. Velocity-dependent offsets in the force curves indicate
\revnew{rate-dependent hysteresis} consistent with the loading-unloading
behavior of PAMs.

\subsubsection{Comparison With Baseline Models}

\begin{figure}[t]
    \centering
    \includegraphics[width=\columnwidth,trim=0 22.1 0 18.7,clip]{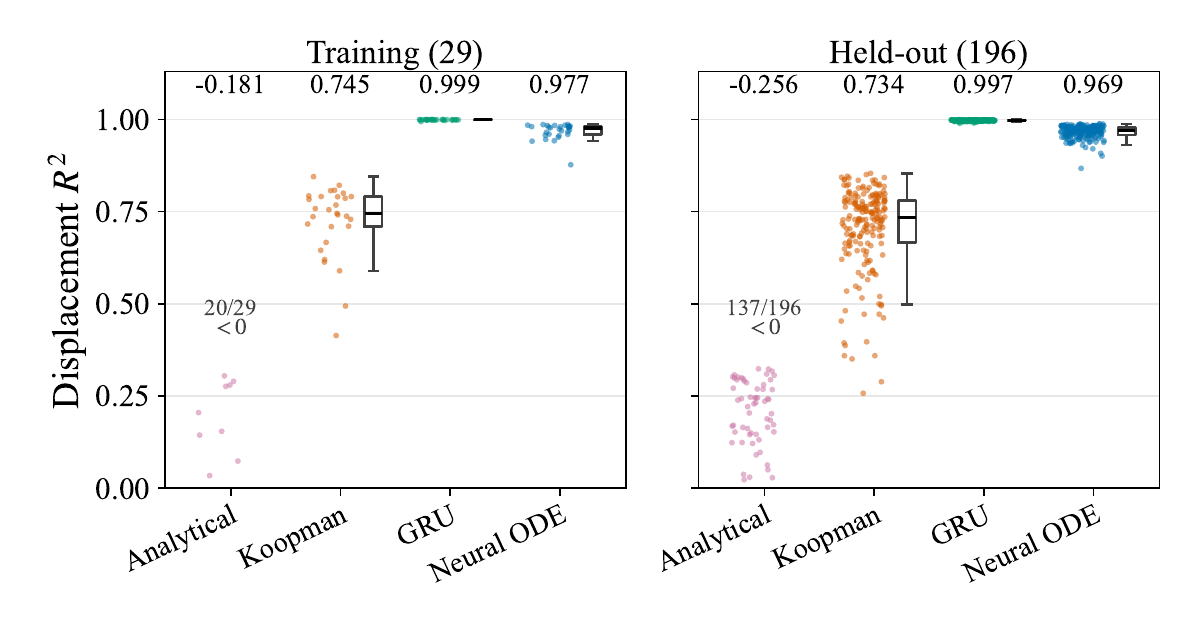}
    \put (-217,109){\textbf{a}}
    \put (-106,109){\textbf{b}}
    \caption{\rev{Displacement-only $R^2$ comparison between the hybrid Neural ODE and
    baseline models. {\textbf{a}} Training datasets. {\textbf{b}} Held-out validation datasets.
    \revnew{Boxes show the median and interquartile range, dots are individual
    conditions, and the numbers above are medians. Negative values (analytical
    model only: 20 of 29 training and 137 of 196 held-out conditions) are not
    plotted.}}}
    \label{fig:comparison}
\end{figure}

\rev{Three displacement-prediction baselines were implemented: a Kang-type
analytical/semi-empirical PAM model \cite{kang2009dynamic}, a Koopman/EDMDc
lifted-regression model \cite{korda2018linear}, and a GRU model
\cite{chung2014empirical} \revnew{(implementation details in
Section~\ref{si:baseline_impl} of the Supplementary Information)}. The
Kang-type model was refitted using single-PAM force--pressure--displacement
data and evaluated with measured pressures, while the Koopman/EDMDc and GRU
models were trained on the same 29 datasets as the proposed model. Because the
Kang-type model requires pressure input, the comparison focuses on
displacement-only $R^2$ over the training datasets
(Fig.~\ref{fig:comparison}{\textbf{a}}) and 196 held-out datasets
(Fig.~\ref{fig:comparison}{\textbf{b}})}.

\rev{\revnew{Table~\ref{tab:model_class_comparison}} summarizes the model
roles. The Kang-type model fits single-PAM forces well but yields negative
$R^2$ in joint-level simulation: the net joint force is a small difference of
large muscle forces, amplifying per-muscle errors. \revnew{The GRU scores
highest but requires 100 measured samples to warm-start, whereas the Neural ODE
uses a single initial state. Like Koopman/EDMDc, the GRU lacks the force and
stiffness representation needed for synthesis. The GRU predicted chamber
pressures more accurately in most held-out conditions (median $R^2$ of 0.97 and
0.96 versus 0.84 and 0.82 for the Neural ODE), but yielded negative $R^2$ for
at least one chamber in 18 of 196 conditions, versus none for the Neural ODE.
Per-state results and example trajectories appear in Fig.~\ref{fig:si_full_r2},
Table~\ref{tab:si_full_r2}, and Fig.~\ref{fig:supp_displacement_comparison}.}}

\begin{figure*}[t]
    \centering
    \includegraphics[width=0.33\textwidth,trim=0 3.9 0 11.7,clip]{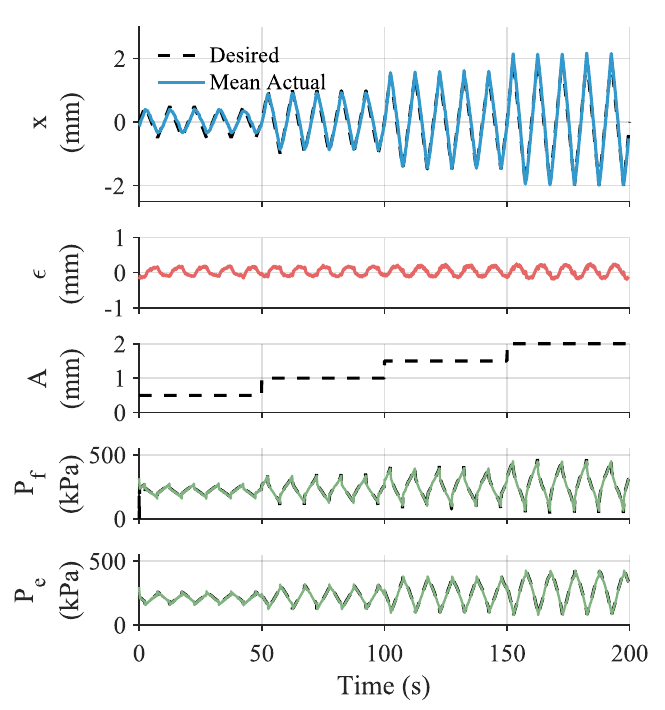}%
    \hfill
    \includegraphics[width=0.33\textwidth,trim=0 3.9 0 11.7,clip]{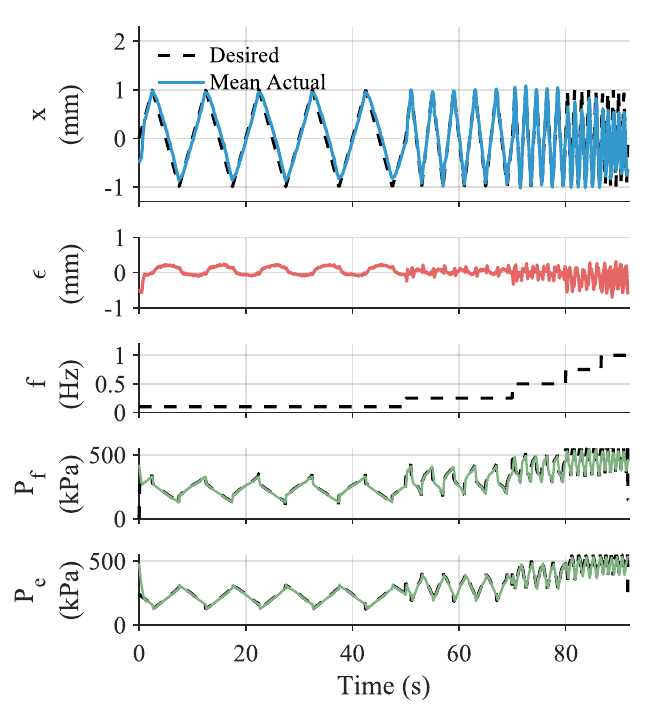}%
    \hfill
    \includegraphics[width=0.33\textwidth,trim=0 3.9 0 11.7,clip]{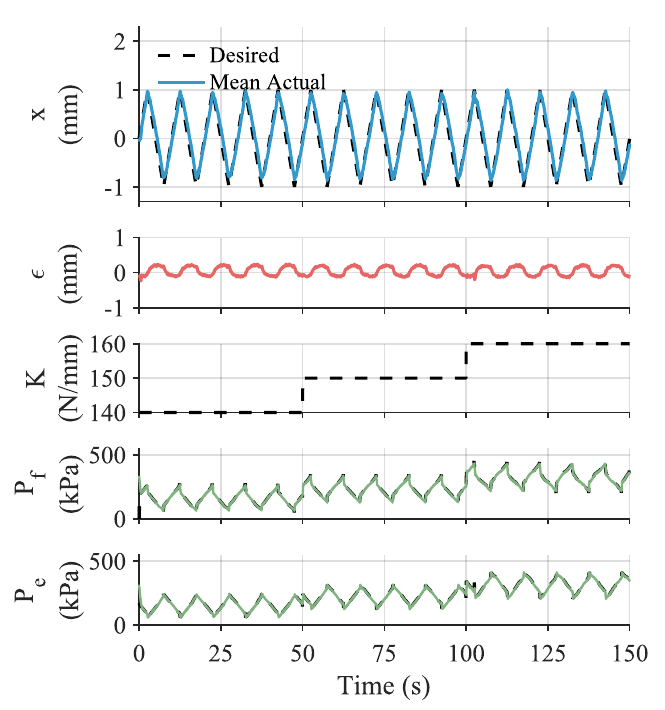}
    % panel labels
    \put (-500,170){\textbf{a}} 
    \put (-330,170){\textbf{b}} 
    \put (-155,170){\textbf{c}}
    \caption{%
    \revnew{Feedforward tracking validation for triangular trajectories under
    {\textbf{a}} amplitude variation ($A=\{0.5,1.0,1.5,2.0\}$~mm),
    {\textbf{b}} frequency variation ($f=\{0.1,0.25,0.5,0.75,1.0\}$~Hz), and
    {\textbf{c}} stiffness variation ($K_d=\{140,150,160\}$~N/mm). Traces are means
    of six repetitions. Sinusoidal trajectories under the same conditions are
    shown in Fig.~\ref{fig:si_tracking_sine}.}}
    \label{fig:tracking_performance}
\end{figure*}

\begin{figure*}[t]
    \centering
    \begin{minipage}[t]{0.48\textwidth}
        \centering
        \includegraphics[width=\linewidth,trim=0 3.8 0 11.0,clip]{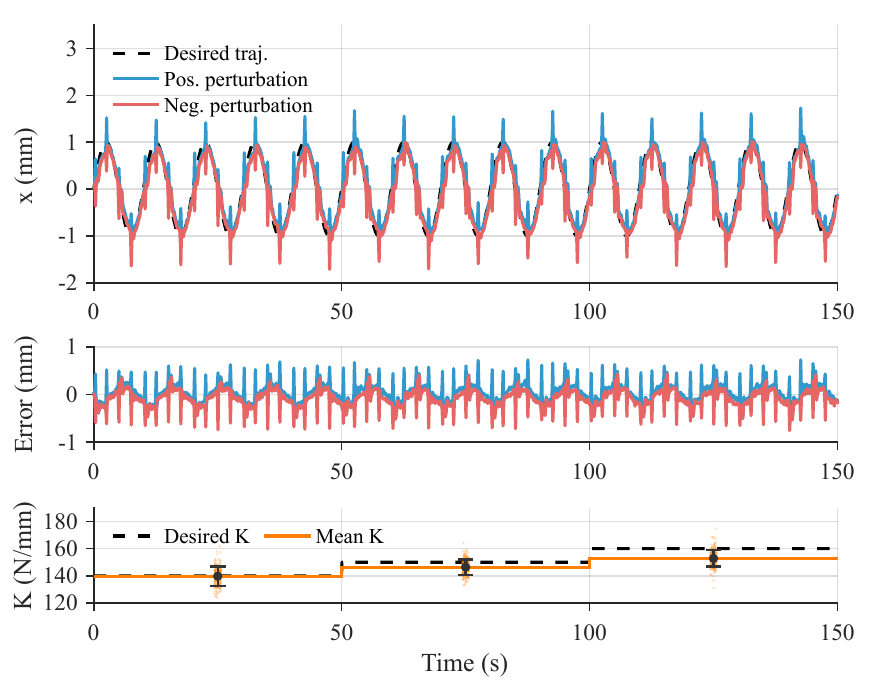}
    \end{minipage}%
    \hfill
    \begin{minipage}[t]{0.48\textwidth}
        \centering
        \includegraphics[width=\linewidth,trim=0 3.8 0 11.0,clip]{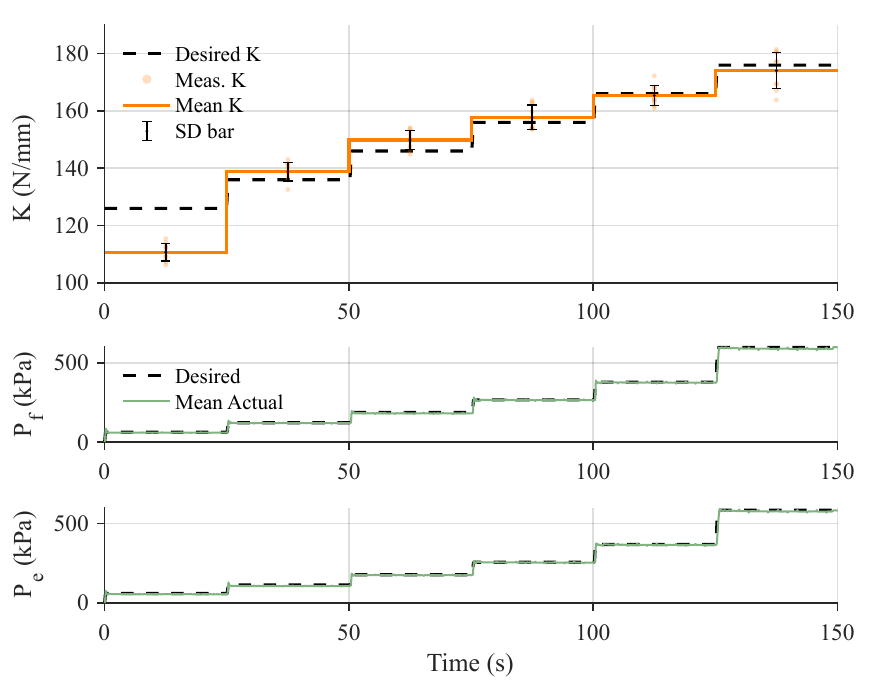}
    \end{minipage}
    \put (-510,177){\textbf{a}}
    \put (-250,177){\textbf{b}}
    \caption{%
    Stiffness synthesis validation.
    {\textbf{a}} Time-varying stiffness tracking during sinusoidal motion with
    $K_d\in\{140,150,160\}$~N/mm; stiffness is identified from closed-valve
    positive and negative perturbation responses at $x_0\in\{0,+1,-1\}$~mm.
    {\textbf{b}} Stiffness range verification at equilibrium for
    \revnew{$K_d=126$--$176$~N/mm}, showing identified stiffness \revnew{and flexor/extensor
    pressures $P_f$ and $P_e$. Identification fits for all pulses are in
    Fig.~\ref{fig:si_sysid}.}}
    \label{fig:stiffness_verification}
\end{figure*}

\subsection{Motion Synthesis Validation}
\label{subsec:results_inverse}

\rev{\revnew{Inverse synthesis runs offline; during experiments, the pressure
controller tracks precomputed commands without online Neural ODE integration or
optimization. On Apple M3 Ultra / AMD EPYC 7713 CPUs, a 70-s forward rollout
takes 0.5/1.1~s and synthesis takes 49/178~ms per point, yielding a 5000-point
table in 4--15~min (Table~\ref{tab:compute}).} We evaluate amplitude,
frequency, and stiffness variations for sinusoidal and triangular
trajectories.}

\subsubsection{Amplitude variation}

At fixed frequency $f = 0.1~\mathrm{Hz}$ and desired stiffness $K_d =
150~\mathrm{N/mm}$, \rev{sinusoidal and triangular trajectories} with
amplitudes $A \in \{0.5,\,1.0,\,1.5,\,2.0\}~\mathrm{mm}$ were commanded. The
resulting tracking errors are summarized in
\revnew{Figs.~\ref{fig:tracking_performance}{\textbf{a}} and
\ref{fig:si_tracking_sine}{\textbf{a}}}. Error magnitude remains nearly
constant across amplitudes, indicating that deviations are dominated by static
friction effects near low-velocity motion reversals rather than by
amplitude-dependent dynamics.

\subsubsection{Frequency variation}

With amplitude fixed at $A = 1.0\,\mathrm{mm}$ and stiffness $K_d =
150\,\mathrm{N/mm}$, \rev{sinusoidal and triangular trajectories} were
commanded at $f \in \{0.1,\,0.25,\,0.5,\,0.75,\,1.0\}\,\mathrm{Hz}$. As shown
in \revnew{Figs.~\ref{fig:tracking_performance}{\textbf{b}} and
\ref{fig:si_tracking_sine}{\textbf{b}}}, the mean tracking error increases with
frequency. \revnew{At 1~Hz, flexor and extensor pressure amplitudes reached
81\% and 96\% of commands, lagging by 28 and 11~ms (vs. 96\% and 99\% without
lag at 0.1~Hz). Higher frequencies also increase airflow during valve
operation, which the closed-valve model does not capture. These effects were
not isolated. Inverse validation was limited to 1~Hz; simulated effects of
pressure lag and attenuation up to 1.5~Hz appear in
Table~\ref{tab:si_simfreq}.}

\subsubsection{Stiffness variation}

To evaluate robustness with respect to stiffness modulation, \rev{sinusoidal
and triangular trajectories were commanded while} stiffness was varied among
$K_d \in \{140,\,150,\,160\}~\mathrm{N/mm}$ at fixed amplitude $A =
1.0~\mathrm{mm}$ and frequency $f = 0.1~\mathrm{Hz}$. Tracking errors remain
comparable across stiffness levels
\revnew{(Figs.~\ref{fig:tracking_performance}{\textbf{c}} and
\ref{fig:si_tracking_sine}{\textbf{c}})}, indicating that stiffness modulation
does not degrade motion tracking accuracy under the tested conditions.

\revnew{Tracking RMSE, averaged over six repetitions, was 0.10--0.16~mm for all
amplitude and stiffness conditions. Frequency caused the largest increase:
sinusoidal RMSE rose from 0.07~mm at 0.5~Hz to 0.37~mm at 1~Hz, and triangular
RMSE from 0.08~mm at 0.25~Hz to 0.32~mm at 1~Hz (Table~\ref{tab:si_rmse}).}

\subsection{Stiffness Synthesis Validation}
\label{subsec:stiffness}

To validate the model's stiffness predictions, the effective joint stiffness was
identified experimentally using perturbation-based system identification under
closed-valve conditions.

\subsubsection{System identification and stiffness estimation}

The effective stiffness $K$ and damping $B$ were identified using a linear
spring-mass-damper model,
\[
  M(\ddot{x}-\ddot{x}_0) + B(\dot{x}-\dot{x}_0) + K(x - x_0) = F_{\mathrm{e}},
\]
where \revnew{$M$ was fixed to the value learned} by the hybrid Neural ODE. The
parameters $K$ and $B$ were identified by minimizing the sum of squared
displacement errors between the measured response and the model integrated with
\texttt{ode45}. A two-stage optimization was used: \texttt{patternsearch} for
initial parameter search, followed by \texttt{fmincon} for local refinement.
The external force $F_{\mathrm{e}}$ was derived from the commanded motor
current and synchronized with the measured displacement.

\subsubsection{Perturbation protocol}

During perturbation for stiffness identification, the pressure controller is
paused, and all valves are kept closed, ensuring constant chamber air masses
throughout the identification window. A motor square torque pulse with a
duration of 150~ms is applied to the joint immediately, and the resulting joint
response is used to identify $K$ and $B$. This places the system in the
closed-valve regime ($\dot{m}_i = 0$) that the learned model describes: if the
servo remained active, valve pulses would alter the air masses and invalidate
the model assumptions. Because the perturbation is applied without delay,
identification is performed at the instantaneous operating condition along the
ongoing trajectory.

A periodic trajectory with $A = 1.0~\mathrm{mm}$ and $f = 0.1~\mathrm{Hz}$ was
tracked while the desired stiffness cycled through $K_d \in
\{140,\,150,\,160\}~\mathrm{N/mm}$. Each stiffness level was held for five
cycles, and both positive and negative perturbation directions were applied for
system identification.

\revnew{Fig.}~\ref{fig:stiffness_verification}{\textbf{a}} illustrates the
system response and stiffness validation. Perturbations are introduced at the
peak, center, and trough of the tracking trajectory in both directions. The
identified stiffness values follow the commanded stiffness schedule across the
tested operating conditions.

To validate the stiffness range predicted by the trained model at the
equilibrium position $x_0 = 0~\mathrm{mm}$, desired stiffness values $K_d \in
\{126,\,136,\,146,\,156,\,166,\,176\}~\mathrm{N/mm}$ were commanded
\revnew{(equivalently $K_q \in [5.96,\,8.32]$~N$\cdot$m/rad at the joint)}.
Positive and negative perturbations were applied, and stiffness was identified
using the same procedure. As shown in
Fig.~\ref{fig:stiffness_verification}{\textbf{b}}, the identified stiffness
values span the predicted range. The lowest stiffness level exhibits a larger
deviation from the target value, associated with operation near the lower
co-contraction region, where one actuator loses effective tension and the
system behavior departs from the pressurized dynamics represented in the model. 

\rev{\revnew{Across 60 pulses, mean identified stiffness was within 3.9~N/mm
(2.6\%) of commands from 136 to 176~N/mm, but was 110.7~N/mm (12\% low) at
126~N/mm. Under identical torque pulses, mean peak deflection decreased
monotonically from 0.67 to 0.40~mm across the commanded range
(Fig.~\ref{fig:si_command}).}}

\rev{Stiffness validation was limited to low-frequency, small-amplitude motion.
At higher frequencies or amplitudes, stronger hysteresis, valve bandwidth
limits, and pressure delay are expected to increase error; broader
perturbation-based validation is left for future work.}

\begin{figure}[t]
    \centering
    \includegraphics[width=\columnwidth,trim=0 8.1 0 28.3,clip]{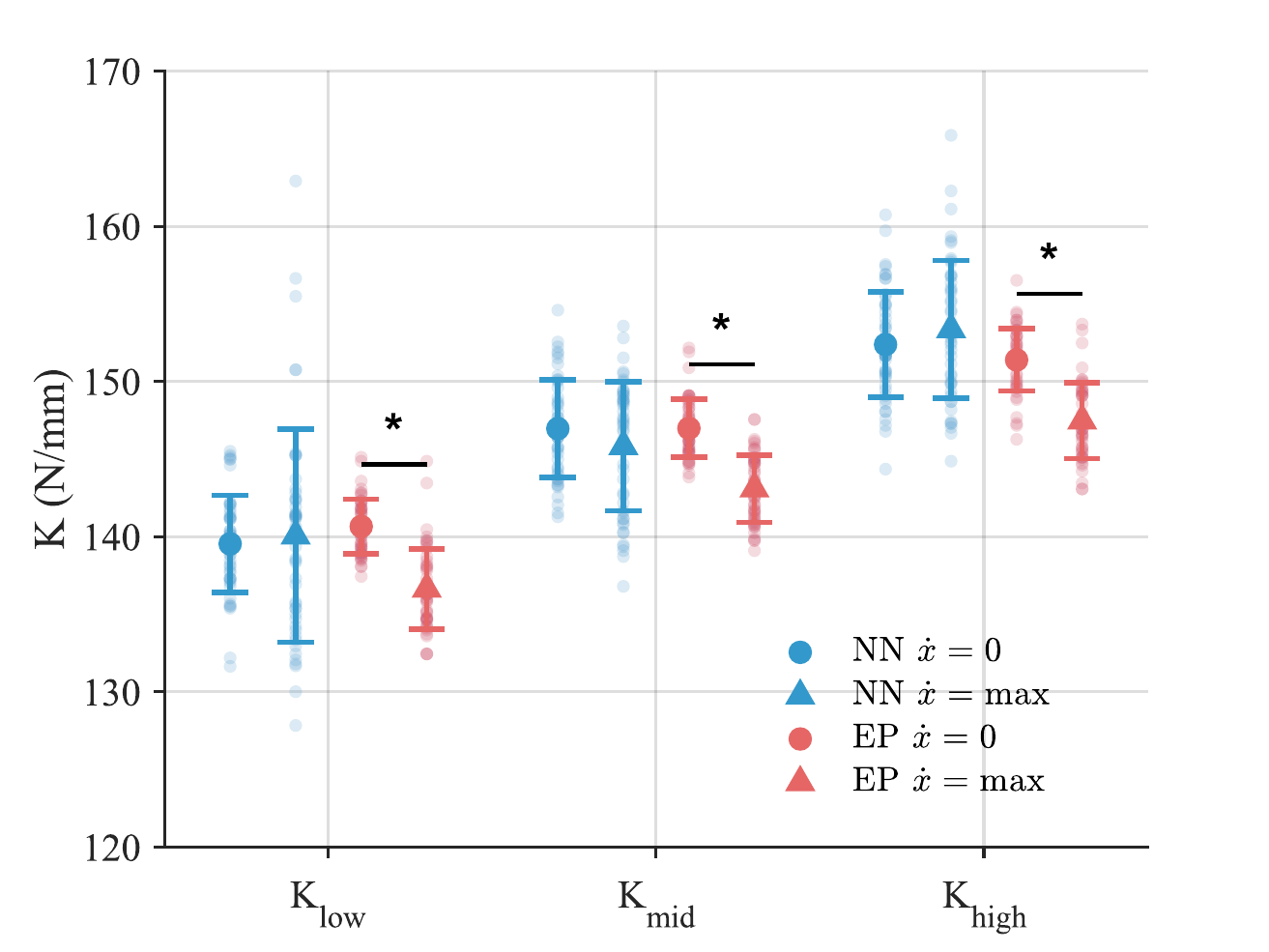}
    \caption{%
    Stiffness comparison between Neural ODE (NN) and equilibrium-point (EP) models.
    Circles and triangles denote perturbations at zero and maximum velocity, respectively;
    large markers show mean $\pm$ standard deviation, and faded markers show individual trials.
    Asterisks indicate significant velocity-dependent stiffness changes for the EP model
    ($p<0.05$).}
    \label{fig:NN_EP_comparison}
\end{figure}

\subsection{Comparison with Equilibrium-Point Model}
\label{subsec:EP_comparison}

To assess whether velocity-dependent force modeling improves stiffness
prediction, the hybrid Neural ODE was compared with an equilibrium-point (EP)
model. The EP model captures the classical antagonistic intuition that
differential pressure governs the equilibrium position, while total pressure
governs stiffness \cite{feldman1986once,ariga2012novel}. Assuming linear
pressure-based mappings, the EP model can be described as
\[
    x_0 = \alpha_1(P_f - P_e) + \alpha_0, \qquad K = \beta_1(P_f + P_e) + \beta_0,
\]
which neglects hysteresis and rate-dependent effects. \rev{To ensure a
controlled comparison, the EP model was fit to experimental
equilibrium-position and stiffness relationships identified from the same 29
excitation datasets used to train the hybrid Neural ODE. The fitted EP model
was then used to generate desired pressure commands, which were evaluated using
the same perturbation-based stiffness-validation procedure applied to the
Neural ODE.}

Both models were evaluated under identical conditions using a sinusoidal
trajectory with stepped stiffness targets. Perturbations were applied at zero
and maximum velocity, and stiffness was identified using the same procedure
described above. \rev{As shown in
\revnew{Table~\ref{tab:stiffness_value_and_percentage}} and
Fig.~\ref{fig:NN_EP_comparison}, the Neural ODE maintains consistent stiffness
across velocities ($\Delta = 1.06$--$2.26\%$, $p \geq 0.1075$), whereas the EP
model exhibits a consistently lower stiffness at maximum velocity and
significantly larger variation ($\Delta = 7.78$--$8.04\%$, paired $t$-test, $p
\leq 3.9\times10^{-12}$).}

The stiffness variation is quantified as
\[
\Delta(\%) =
\frac{\left|K_{\dot{x}=0} - K_{\dot{x}=\dot{x}_{\max}}\right|}
{K_{\max,\mathrm{theory}} - K_{\min,\mathrm{theory}}}
\times 100\%,
\]
where $K_{\min,\mathrm{theory}} = 126$~N/mm and
$K_{\max,\mathrm{theory}} = 176$~N/mm define the theoretical stiffness range.

\subsection{Feedforward Load Holding}
\label{subsec:payload}

\begin{figure}[t]
    \centering
    \includegraphics[width=\columnwidth]{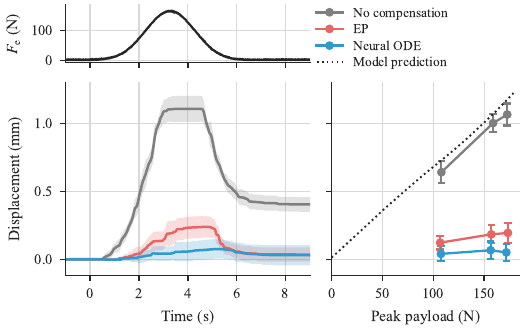}
    \put (-112,146){\textbf{a}}
    \put (-112,109){\textbf{b}}
    \put (-89,109){\textbf{c}}
    \caption{%
    \revnew{Feedforward load holding at a commanded stiffness of 150~N/mm. {\textbf{a}}
    Payload force applied by the motor (165~N commanded, 173~N measured).
    {\textbf{b}} Displacement in the load direction, mean $\pm$ SD of 10 trials.
    {\textbf{c}} Displacement at peak payload against measured peak payload;
    dotted line, model prediction without compensation.}}
    \label{fig:payload}
\end{figure}

\revnew{To test feedforward payload compensation, the joint was held at $x=0$
and 150~N/mm commanded stiffness while the motor applied Gaussian payloads
($\sigma=1$~s; commanded peaks of 100, 150, and 165~N; measured 107, 158, and
173~N; both directions; 10 trials each). Three pressure plans were compared:
the load-free Neural ODE (no compensation), the EP model
(Section~\ref{subsec:EP_comparison}) with payload added to its equilibrium, and
the Neural ODE with payload included in \eqref{eq:synthesis}. Without
compensation, the joint sagged 0.64--1.07~mm, versus 0.73--1.18~mm predicted by
the model's static force balance (7--14\% high), consistent with its held-out
amplitude error (Section~\ref{subsec:forward_model}). A residual 0.3--0.4~mm
displacement after unloading reflected unmodeled history effects. Peak
displacement was 0.12--0.19~mm with EP compensation, increasing with payload,
but only 0.04--0.07~mm with Neural ODE compensation, comparable to
trial-to-trial SD (0.06~mm) (Fig.~\ref{fig:payload},
Tables~\ref{tab:si_payload} and \ref{tab:si_payload_pred}). Because the payload
was known in advance, this validates feedforward planning, not closed-loop
control.}

\section{Conclusion}
\label{sec:conclusion}

This paper presented a hybrid Neural ODE framework for modeling antagonistic
pneumatic artificial muscle dynamics. By embedding parametric joint
mechanics and pressure dynamics into a continuous-time learning formulation,
the approach captures coupled motion, pneumatic state evolution, and nonlinear
antagonistic force behavior in a physically structured yet data-driven manner.

\rev{The forward model predicts joint motion and chamber pressures on held-out
co-contraction conditions, and the learned scalar force model captures
equilibrium shifts, stiffness scaling, and \revnew{rate-dependent} hysteresis.}

A feedforward input synthesis procedure derived from the learned dynamics
\revnew{was used to verify the model experimentally. Tracking RMSE was 0.07--0.16~mm up to
0.5~Hz and 0.32--0.37~mm at 1~Hz. Closed-valve stiffness was within 3\% of
commanded values over 136--176~N/mm but 12\% below the 126~N/mm command.
Velocity-dependent force modeling improved stiffness consistency over an
equilibrium-point model. Under payloads up to 173~N, the model predicted
displacement within 14\% and maintained position within 0.04--0.07~mm via
feedforward compensation. For soft wearables, where pneumatic actuation is
common~\cite{bardi2022upper}, a joint can thus be set compliant while the user
moves it and stiff while it holds a posture or load, in each case with the
valves closed.}

\revnew{The framework is limited to closed-valve dynamics, with a force model
trained at constant air mass and no hysteresis state. Synthesis is offline.
Validation was confined to one platform, known motor-applied loads, and
characterized pressure and frequency ranges, without users or added inertia.
The lookup-table implementation cannot adapt to unmodeled disturbances or
unseen trajectories. Future work will incorporate valve and mass-flow dynamics
for online replanning and user testing.}

\bibliographystyle{IEEEtran}
\bibliography{bib}

% Author biographies: omitted in the arXiv version (as in arXiv v2).
%\begin{IEEEbiography}{Xinyao Wang}
%(Student Member, IEEE) received the BEng degree in automotive engineering from
%Wuhan University of Technology, Wuhan, China, in 2020, and a MS degree in
%mechanical engineering from the University of California, Riverside (UCR),
%Riverside, CA, USA, in 2021, where he is currently pursuing a PhD degree. His
%research interests include modeling and control of pneumatic artificial muscles
%for wearable robotic systems.
%\end{IEEEbiography}

%\begin{IEEEbiography}{Jonathan Realmuto}
%(Member, IEEE) received a PhD\ in mechanical engineering from the University of
%Washington, Seattle, WA, USA, in 2017. He was a Postdoctoral Researcher with
%the University of California, Irvine, CA, USA. He is currently an Assistant
%Professor with the Department of Mechanical Engineering, University of
%California, Riverside, CA, USA. His research interests include soft robotic
%actuators, wearable robots, and learning-based modeling and control of
%pneumatic systems.
%\end{IEEEbiography}

% Single-column layout from 2026-11-11 (V3). v2 opened with \clearpage only.
% The SI is still built by \input from main.tex (so references to main-text
% figures and sections resolve) and its pages are extracted.
\clearpage
\onecolumn
\setcounter{page}{1}
\renewcommand{\thepage}{S\arabic{page}}
\section*{Supplementary Information for: \\
``Physics-Embedded Neural ODEs for Learning
Antagonistic Pneumatic Artificial Muscle Dynamics'' \\
{\normalfont\normalsize Xinyao Wang and Jonathan Realmuto}}

\setcounter{figure}{0}
\renewcommand{\thefigure}{S\arabic{figure}}
\setcounter{table}{0}
\renewcommand{\thetable}{S\arabic{table}}
\providecommand{\theHfigure}{}\renewcommand{\theHfigure}{S\arabic{figure}}
\providecommand{\theHtable}{}\renewcommand{\theHtable}{S\arabic{table}}
\newcounter{sisec}
\renewcommand{\thesisec}{\Alph{sisec}}
\providecommand{\theHsisec}{}\renewcommand{\theHsisec}{SI.\Alph{sisec}}
\newcommand{\sisection}[1]{\refstepcounter{sisec}\subsection*{\thesisec. #1}}
\providecommand{\revnew}{}\renewcommand{\revnew}[1]{#1}
\noindent\textbf{Contents.} The sections follow the order in which the main text cites them.
\begin{center}
\small
\begin{tabular}{@{}clllc@{}}
\toprule
 & Section & Figures and tables & Cited in Section & Page \\
\midrule
\ref{si:prior} & Relation to Prior Work & Table~\ref{tab:si_prior} & I & \pageref{si:prior} \\
\ref{si:eps} & Sensitivity to the Synthesis Tolerance & Fig.~\ref{fig:si_eps} & \ref{subsec:inverse_model} & \pageref{si:eps} \\
\ref{si:curriculum} & Training Curriculum & Table~\ref{tab:curriculum} & \ref{subsec:stagetrain} & \pageref{si:curriculum} \\
\ref{si:compute} & Computational Cost & Table~\ref{tab:compute} & \ref{subsec:stagetrain}, \ref{subsec:results_inverse} & \pageref{si:compute} \\
\ref{si:heldout} & Forward Model Accuracy Across Held-Out Conditions & Figs.~\ref{fig:si_r2_hist}, \ref{fig:si_best_worst}; Table~\ref{tab:si_loop} & \ref{subsec:forward_model} & \pageref{si:heldout} \\
\ref{si:pulses} & Prediction of Closed-Valve Torque Pulses & Fig.~\ref{fig:si_pulse} & \ref{subsec:forward_model} & \pageref{si:pulses} \\
\ref{si:baseline_impl} & Baseline Model Implementation Details & --- & \ref{subsec:forward_model} & \pageref{si:baseline_impl} \\
\ref{si:baseline_cmp} & Baseline Comparison & Tables~\ref{tab:model_class_comparison}, \ref{tab:si_full_r2}; Figs.~\ref{fig:si_full_r2}, \ref{fig:supp_displacement_comparison} & \ref{subsec:forward_model} & \pageref{si:baseline_cmp} \\
\ref{si:tracking_sine} & Tracking of Sinusoidal Trajectories & Fig.~\ref{fig:si_tracking_sine} & \ref{subsec:results_inverse} & \pageref{si:tracking_sine} \\
\ref{si:simfreq} & Simulated Tracking Error Above 1~Hz & Table~\ref{tab:si_simfreq} & \ref{subsec:results_inverse} & \pageref{si:simfreq} \\
\ref{si:rmse} & Tracking Error by Condition & Table~\ref{tab:si_rmse} & \ref{subsec:results_inverse} & \pageref{si:rmse} \\
\ref{si:command} & Commanded and Identified Stiffness & Figs.~\ref{fig:si_sysid}, \ref{fig:si_command} & \ref{subsec:stiffness} & \pageref{si:command} \\
\ref{si:ep} & Comparison with the Equilibrium-Point Model & Table~\ref{tab:stiffness_value_and_percentage} & \ref{subsec:EP_comparison} & \pageref{si:ep} \\
\ref{si:payload} & Feedforward Load Holding & Tables~\ref{tab:si_payload}, \ref{tab:si_payload_pred} & \ref{subsec:payload} & \pageref{si:payload} \\
\bottomrule
\end{tabular}
\end{center}

\sisection{Relation to Prior Work}\label{si:prior}

\revnew{Table~\ref{tab:si_prior} compares this work with the closest prior
work on models and controllers for antagonistic pneumatic joints and on Neural
ODE models of soft robots. Reference numbers are those of the main text.}

\begin{table}[!htb]
\centering
\caption{Closest prior work and this work.}
\label{tab:si_prior}
\footnotesize
\begin{tabular}{@{}p{0.13\linewidth}p{0.15\linewidth}p{0.21\linewidth}p{0.17\linewidth}p{0.11\linewidth}p{0.10\linewidth}@{}}
\toprule
Work & System & Model & Joint stiffness in the model & Stiffness measurements needed to build it & Stiffness identified independently during motion \\
\midrule
Trumi\'c et al.\ \cite{trumic2021decoupled} & Antagonistic McKibben joints, hardware & Static maps from valve voltage to joint torque and stiffness, fitted to steady-state tests; adaptive feedback controller & Computed from the fitted static map & No (checked with static weights) & No \\
Habich et al.\ \cite{habich2023learning} & Antagonistic soft pneumatic actuator, test bench & Gaussian processes from desired angle and stiffness to the two pressures & Target of the map, identified beforehand & Yes, at 518 static pressure pairs & No \\
Bachhuber et al.\ \cite{bachhuber2024soft} & Antagonistic PAM joint, hardware & Neural ODE with nine latent states, from pressure difference to joint angle at one mean pressure & Not represented & --- & --- \\
Takeda et al.\ \cite{takeda2025encrypted} & Antagonistic PAM joint, hardware & Polynomial static force model in pressure and angle; PI feedback & Angle derivative of the fitted torque & No & No \\
Yan et al.\ \cite{yan2025robust} & Antagonistic PAM platform, hardware & Three-element PAM model with a neural-network compensator; predictive force control & Not predicted or commanded & --- & --- \\
Youm and Kim \cite{youm2026physics} & Multi-segment soft robot, simulation & Neural ODE trained with an Euler--Lagrange residual loss & Structural coefficient, identified as a parameter; not commanded & --- & --- \\
\midrule
This work & Antagonistic PAM joint, hardware & Neural ODE with physical states; joint and gas equations explicit, net force learned & Position derivative of the learned force at constant air mass & No & Yes, from closed-valve pulses during tracking \\
\bottomrule
\end{tabular}
\end{table}

\sisection{Sensitivity to the Synthesis Tolerance}\label{si:eps}

\begin{figure}[!htb]
    \centering
    \includegraphics[width=0.75\textwidth]{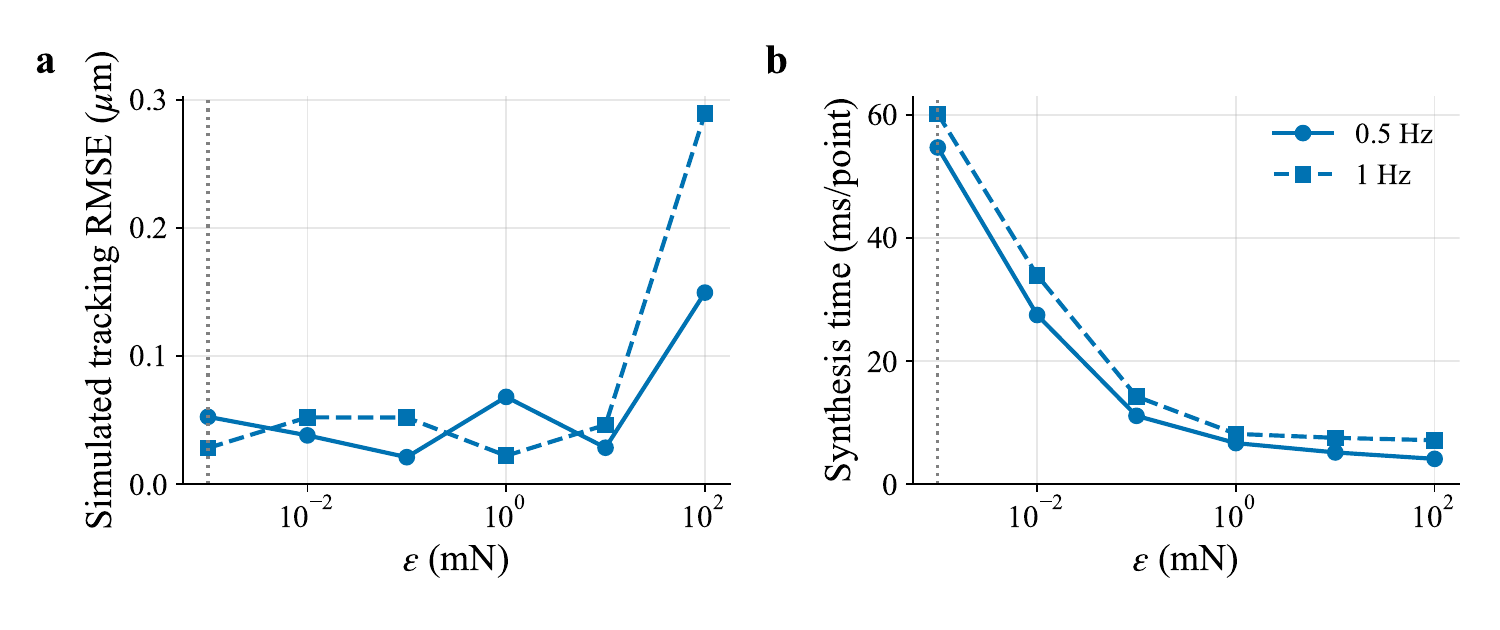}
    \caption{%
    Effect of the synthesis tolerance \(\varepsilon\) for a 1-mm sinusoid at
    0.5 and 1~Hz (\(K_d=150\)~N/mm). \textbf{a} Tracking RMSE when the model is
    driven by the synthesized air masses. \textbf{b} Mean synthesis time per
    lookup-table point (Apple M3 Ultra). The dotted line marks
    \(\varepsilon=10^{-3}\)~mN, the value used in this work.}
    \label{fig:si_eps}
\end{figure}

\revnew{The synthesis of Section~\ref{subsec:inverse_model} was repeated with
the tolerance $\varepsilon$ varied from $10^{-3}$ to $10^{2}$~mN, and tracking
was simulated by driving the model with the synthesized air masses
(Fig.~\ref{fig:si_eps}). The simulated tracking error stayed below
0.3~$\mu$m at every tolerance, while the mean synthesis time per point fell
from 55--60~ms to 4--8~ms.}

\sisection{Training Curriculum}\label{si:curriculum}

\revnew{Table~\ref{tab:curriculum} lists the training conditions of each stage of the
curriculum of Section~\ref{subsec:stagetrain}.}

\begin{table}[!htb]
\centering
\caption{Staged training curriculum. Each stage starts from the best checkpoint
of the previous stage; $B$ is the prescribed auxiliary damping. Dataset
conditions are reported as rounded absolute pressures in kPa and were originally
defined using gauge-pressure labels in 5-psi increments.}
\label{tab:curriculum}
\small
\setlength{\tabcolsep}{2pt}
\begin{tabular}{c c l}
\hline
Stage & $B$ (kg/s) & Training datasets (kPa) \\
\hline
1--7 & $6500 \rightarrow 0$ & 170-170 \\ 
8 & 0 & 170-170, 412-412, 653-653 \\ 
9 & 0 & previous + 377-446, 446-377 \\ 
10 & 0 & previous + 308-515, 515-308 \\ 
11 & 0 & previous + 239-584, 584-239 \\ 
12 & 0 & previous + 170-653, 653-170 \\ 
13 & 0 & previous + 412-308, 412-515, 308-412, 515-412 \\ 
14 & 0 & previous + 412-239, 412-584, 239-412, 584-412 \\ 
15 & 0 & previous + 412-170, 412-653, 170-412, 653-412 \\ 
16 & 0 & previous + 239-239, 308-308, 377-377, 446-446, 515-515, 584-584 \\
\hline
\end{tabular}
\end{table}

\sisection{Computational Cost}\label{si:compute}

\revnew{Table~\ref{tab:compute} compares the training and rollout cost of the
hybrid Neural ODE and the baselines, each timed with the weights used for the
reported results.}

\revnew{Training is slow for three reasons. First, every optimizer step
integrates each training recording in full, about $7\times10^{4}$ fixed 1-ms
steps of the fifth-order solver, and then differentiates backward through all
of them. The adjoint sensitivities are computed in discrete form, by
reverse-mode differentiation through the solver steps with recursive
checkpointing; this gives the exact gradient of the discretized model, whereas
integrating a continuous adjoint equation gives an approximate one. The cost
therefore scales with the number of solver steps. Second, the steps are
sequential in time and each is small (four states and a network with two hidden
layers of 96 units), so the computation is essentially serial: on the 8-core
node the average use was about 1.3 cores. Third, the published code integrates
the recordings one after another inside one compiled function, so compilation
time and memory grow with the number of recordings (1.7~h and 93~GiB with 29).}

\revnew{Two changes reduce the cost without changing the method. Stacking the
recordings of a step into a single integration with a batched state gives the
same losses (within $10^{-4}$, relative) and gradients (within $10^{-3}$), with 7 times less time
per step, 5.5~min of compilation and 3.6~GiB of memory at 29 recordings. Over
the whole curriculum the gain is smaller (127~h against 333~h), because most
steps use one to seven recordings. Replacing the solver with Heun's second-order
method at 2-ms steps cuts the force evaluations per unit of simulated time about
sixfold and changed the loss by about 1\% over six test steps. With both changes
the curriculum takes about 25~h on the EPYC node (projected) and 10.3~h on the
M3 Ultra (measured).}

\begin{table}[!htb]
\centering
\caption{Computational cost of the hybrid Neural ODE and the baselines, timed on one CPU node of a cluster (AMD EPYC 7713, 8 cores) and on a workstation (Apple M3 Ultra, 28 cores, 96~GB). Rollout: one 70-s held-out trajectory (about $7\times10^{4}$ steps of 1~ms), median of 3 runs on 3 held-out conditions. Training is a one-time, offline cost; no model is trained or optimized during execution.}
\label{tab:compute}
\begin{threeparttable}
\small
\begin{tabular}{lcccccc}
\toprule
 & \multicolumn{2}{c}{Training} & \multicolumn{2}{c}{Rollout (70 s)} & \multicolumn{2}{c}{Per 1-ms step} \\
\cmidrule(lr){2-3}\cmidrule(lr){4-5}\cmidrule(lr){6-7}
Model & EPYC & M3 Ultra & EPYC & M3 Ultra & EPYC & M3 Ultra \\
\midrule
Kang-type$^{a}$     & 25~s             & 14.5~s           & 1.5~s       & 0.58~s       & 21~$\mu$s & 8~$\mu$s \\
Koopman/EDMDc$^{a}$ & 0.5~s            & 0.23~s           & 1.5~s       & 0.56~s       & 22~$\mu$s & 8~$\mu$s \\
GRU                 & 6--15~h$^{b}$    & 3--8~h$^{b}$     & 0.7~s$^{c}$ & 0.37~s$^{c}$ & 10~$\mu$s & 5~$\mu$s \\
Neural ODE          & $\approx$333~h$^{d}$ & $\geq$108~h$^{d}$ & 1.1~s$^{e}$ & 0.47~s$^{e}$ & 16~$\mu$s & 7~$\mu$s \\
Neural ODE, batched & $\approx$127~h$^{g}$ & $\approx$45~h$^{g}$ & --- & --- & --- & --- \\
Neural ODE, batched + Heun 2~ms & $\approx$25~h$^{h}$ & 10.3~h$^{h}$ & --- & --- & --- & --- \\
\midrule
 & & & \multicolumn{2}{c}{Synthesis, per point$^{f}$} & & \\
\cmidrule(lr){4-5}
Neural ODE          & ---              & ---              & 178~ms      & 49~ms        &           &          \\
\bottomrule
\end{tabular}
\begin{tablenotes}
\footnotesize
\item[$^{a}$] Single-muscle coefficient fit (Kang-type); one least-squares solve (Koopman/EDMDc).
\item[$^{b}$] 12--32 epochs (best checkpoint at epoch 12) at 28~min (EPYC) and 15~min (M3 Ultra) per epoch; 7.7~min per epoch on an NVIDIA A100.
\item[$^{c}$] Compiled re-implementation of the published GRU (identical $R^2$); the reference implementation, stepped in Python, takes 17~s (EPYC) and 3.0~s (M3 Ultra).
\item[$^{d}$] About $7.0\times10^{4}$ dataset-epochs across 16 curriculum stages (12,581 steps), launched manually. EPYC: about 333~h, from per-dataset-epoch times of 15.8~s (one dataset) and 18.5~s (29 datasets) interpolated over the schedule, plus about 1.7~h compilation and about 93~GiB peak memory. M3 Ultra: at least 108~h, from 5.6~s per dataset-epoch timed on one dataset. Both exclude epochs after each stage's best checkpoint, which were not logged.
\item[$^{e}$] After a one-time compilation of 25~s (EPYC) and 10~s (M3 Ultra).
\item[$^{f}$] Inverse synthesis of one lookup-table point, with the objective and its gradient compiled.
\item[$^{g}$] Same curriculum, step counts and solver (Tsit5, 1~ms) as the published code, with all datasets of a step in one batched solve. Losses agree to within $10^{-4}$ and gradients to within $10^{-3}$ (relative). EPYC and M3 Ultra values are projections from measured per-step times.
\item[$^{h}$] Batched, with 2-ms Heun integration. The EPYC value is projected from measured per-step times (23.6~h of training and 1.2~h of compilation); the M3 Ultra value is measured. A model trained this way reached held-out trajectory accuracy similar to the published model; it is not used for the results in this paper.
\end{tablenotes}
\end{threeparttable}
\end{table}

\sisection{Forward Model Accuracy Across Held-Out Conditions}\label{si:heldout}

\begin{figure}[!htb]
    \centering
    \includegraphics[width=0.5\textwidth]{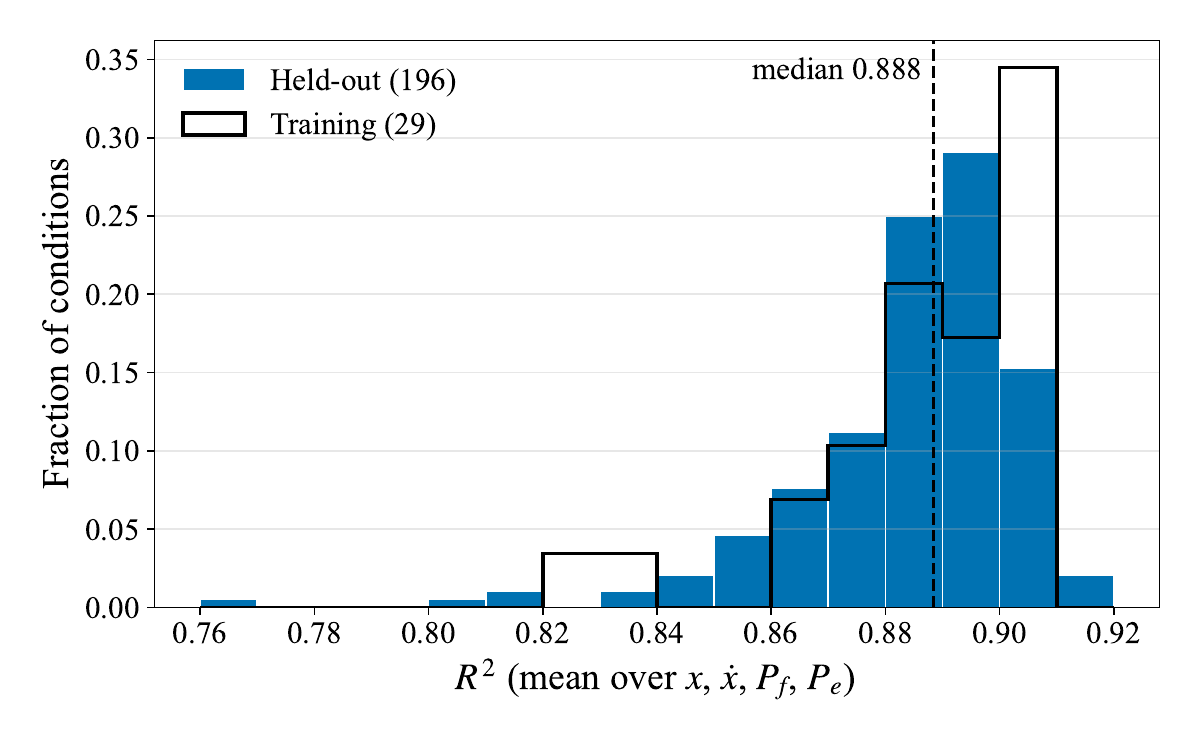}
    \caption{%
    Distribution of the Neural ODE \(R^2\) (mean over \(x\), \(\dot{x}\),
    \(P_f\), \(P_e\)) over the 196 held-out conditions (bars) and the 29
    training conditions (outline). The dashed line marks the held-out median.}
    \label{fig:si_r2_hist}
\end{figure}

\begin{figure}[!htb]
    \centering
    \makebox[\textwidth][l]{\textbf{a}}\\[-1ex]
    \includegraphics[width=0.92\textwidth]{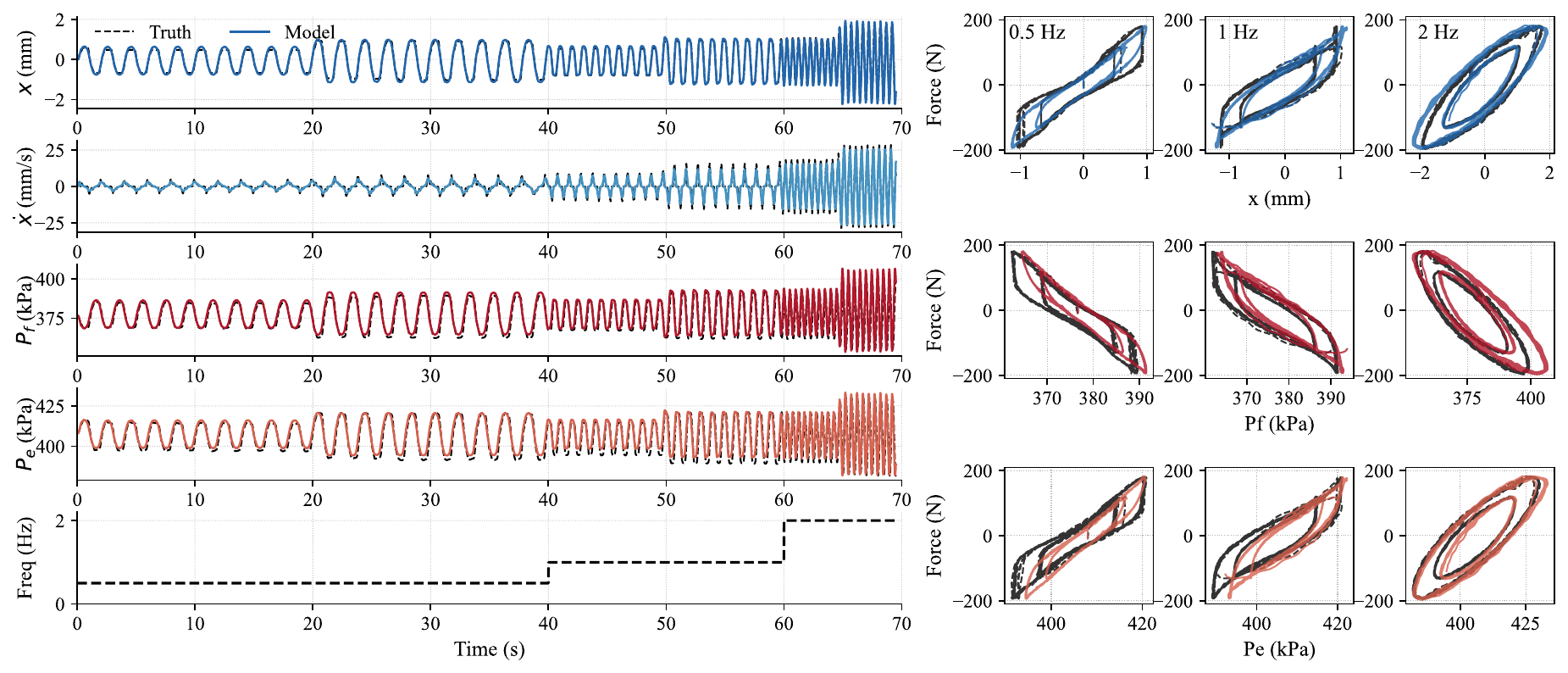}\\[1ex]
    \makebox[\textwidth][l]{\textbf{b}}\\[-1ex]
    \includegraphics[width=0.92\textwidth]{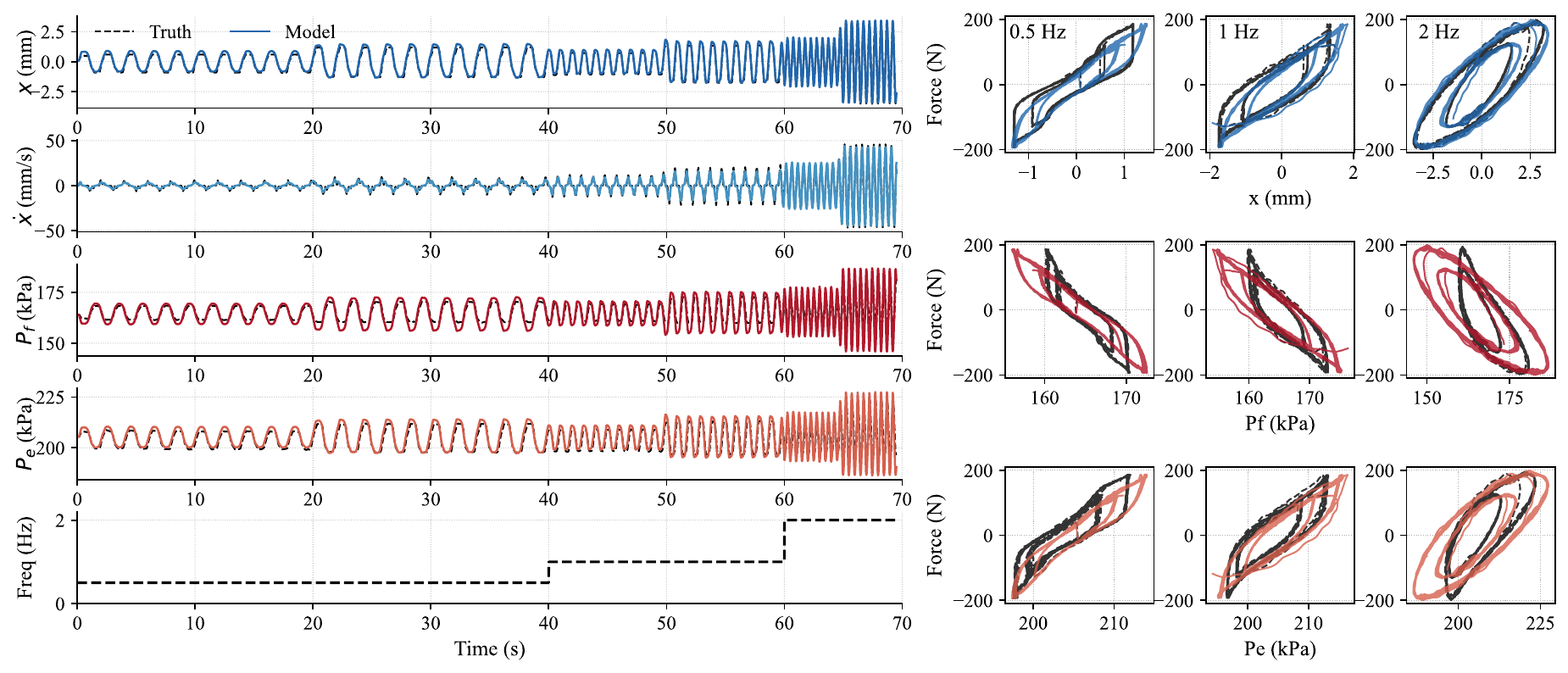}
    \caption{%
    Best (\textbf{a}, 40-45~psi gauge, four-state mean \(R^2=0.913\)) and worst
    (\textbf{b}, 10-15~psi gauge, \(R^2=0.768\)) of the 196 held-out
    conditions, from the same model and in the format of
    Fig.~\ref{fig:response_example}. In the worst case the displacement is
    still predicted (\(R^2=0.973\)), but the measured flexor pressure varies
    less than predicted (\(R^2=0.39\)), consistent with the slack-tendon
    behavior at low co-contraction described in
    Section~\ref{subsec:forward_model}.}
    \label{fig:si_best_worst}
\end{figure}

\revnew{Fig.~\ref{fig:si_r2_hist} shows the distribution of $R^2$, averaged
over the four states, across the 196 held-out conditions (mean 0.885, median
0.888, range 0.768--0.913). Fig.~\ref{fig:si_best_worst} shows the best and
worst held-out conditions in the format of Fig.~\ref{fig:response_example}.}

\revnew{Table~\ref{tab:si_loop} separates the held-out results by excitation
frequency and load amplitude. With constant air masses, the area of the
force--displacement loop is the work dissipated per cycle. Both load
amplitudes are present in the training recordings, so the load itself is not
a held-out variable.}

\begin{table}[!htb]
\centering
\caption{Held-out accuracy by excitation frequency and load amplitude (196 held-out conditions per row). $A$ is the displacement amplitude and $W$ the area of the force--displacement loop (work dissipated per cycle); ratios are predicted over measured, given as median [interquartile range].}
\label{tab:si_loop}
\begin{threeparttable}
\small
\begin{tabular}{cccccc}
\toprule
Frequency & Load & $R^2_x$ & $A$ ratio & Measured $W$ & $W$ ratio \\
(Hz) & (N) & median (min) & & (mJ) & \\
\midrule
0.5 & 126 & 0.941 (0.52) & 1.09 [1.05--1.12] & 79 & 0.88 [0.85--0.93] \\
0.5 & 188 & 0.967 (0.87) & 1.12 [1.09--1.16] & 166 & 0.81 [0.77--0.84] \\
\midrule
1 & 125 & 0.976 (0.82) & 1.11 [1.08--1.15] & 94 & 0.99 [0.96--1.03] \\
1 & 187 & 0.979 (0.92) & 1.17 [1.14--1.20] & 199 & 0.95 [0.93--0.98] \\
\midrule
2 & 126 & 0.989 (0.94) & 1.00 [0.99--1.03] & 225 & 1.00 [0.97--1.06] \\
2 & 190 & 0.981 (0.91) & 1.09 [1.04--1.17] & 507 & 1.09 [0.99--1.21] \\
\bottomrule
\end{tabular}
\begin{tablenotes}
\footnotesize
\item Both load amplitudes are present in the recordings of the training conditions. Measured $W$ is the median over conditions.
\end{tablenotes}
\end{threeparttable}
\end{table}

\sisection{Prediction of Closed-Valve Torque Pulses}\label{si:pulses}

\begin{figure}[!htb]
    \centering
    \includegraphics[width=\textwidth]{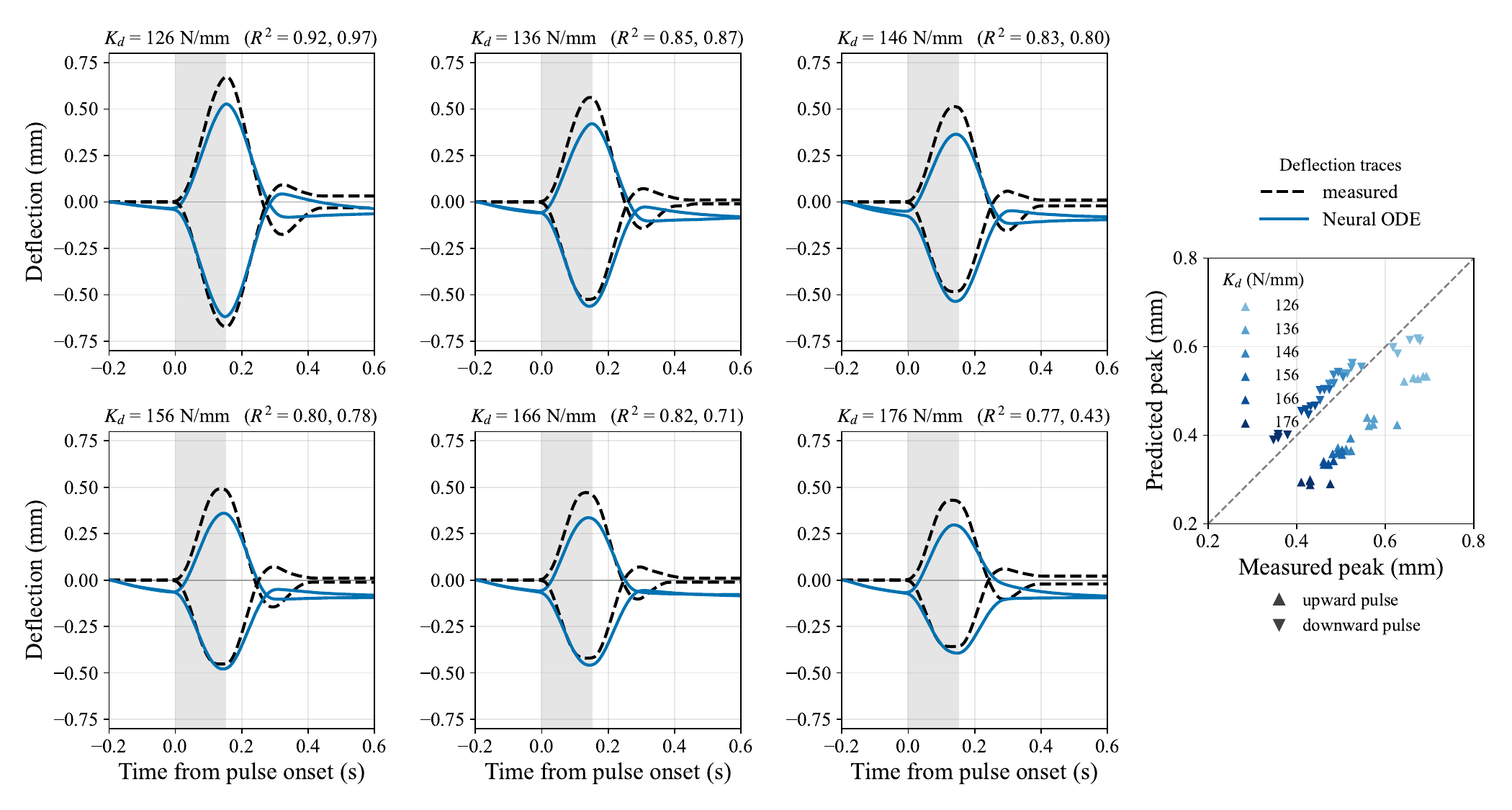}
    \put (-503,260){\textbf{a}}
    \put (-126,188){\textbf{b}}
    \caption{%
    Neural ODE prediction of closed-valve torque pulses, an input not used in
    training. \textbf{a} Measured and predicted deflection for two pulses at each
    commanded stiffness \(K_d\), one in each direction (the pulse with the
    median \(R^2\) of its direction; the two \(R^2\) values are given above
    each panel; shaded: the 150-ms pulse). \textbf{b}
    Predicted versus measured peak deflection for all 60 pulses; upward- and
    downward-pointing triangles mark pulses that deflect the joint upward and
    downward in \textbf{a}.}
    \label{fig:si_pulse}
\end{figure}

\revnew{The 60 torque pulses of the stiffness-range experiment in
Section~\ref{subsec:stiffness} (150~ms, about 64~N, valves closed) are an
input that was not used in training. For each pulse, the model was started
from the measured state 0.2~s before pulse onset, with air masses estimated as
in Section~\ref{subsec:mass_estimation}, and driven by the measured force
(Fig.~\ref{fig:si_pulse}). The median displacement $R^2$ was 0.81 (minimum
0.40). The predicted peak deflection was 12\% below the measured peak on
average: 27\% low for pulses in one direction and 4\% high in the other. The
rest position of the model, set by the estimated air masses, differed from the
measured one by about 0.08~mm on average. This appears as a drift of the
predicted trace before pulse onset and accounts for the difference between the
two directions: the mean peak errors of $-0.14$ and $+0.02$~mm are a common
under-prediction of 0.06~mm shifted by $\mp0.08$~mm.}

\sisection{Baseline Model Implementation Details}\label{si:baseline_impl}

All three baselines were evaluated for displacement prediction using the same
training and held-out dataset split as the main comparison in Section~\ref{subsec:forward_model}.

\subsubsection*{1) Kang-type analytical/semi-empirical PAM model}

The Kang-type baseline follows the analytical/semi-empirical PAM force structure
used for McKibben muscles. The model was first fitted using single-PAM
force--pressure--displacement data. The fitted force model is written as
\[
F_{\mathrm{PAM}}
=
A_0 p_g
\left[
\frac{3(1-q\epsilon)^2}{\tan^2(\alpha_0)}
-
\frac{1}{\sin^2(\alpha_0)}
\right]
-
c_v \dot{\epsilon}
-
c_c \operatorname{sgn}(\dot{\epsilon}),
\]
where \(p_g\) is gauge pressure, \(\epsilon\) is normalized contraction,
\(\dot{\epsilon}\) is contraction rate, \(A_0\) is the nominal cross-sectional
area, and \(\alpha_0\) is the fixed initial braid angle. The pressure-dependent
correction factor is
\[
q = 1 + c_{q1}\exp(c_{q2}p_g).
\]
The fitted parameters are \(c_{q1}\), \(c_{q2}\), \(c_v\), and \(c_c\).
In this implementation, \(D_0=10~\mathrm{mm}\), \(L_0=200~\mathrm{mm}\), and
\(\alpha_0=27.5^\circ\) were fixed from actuator geometry. After
fitting the single-PAM force model, the antagonistic joint was simulated using
the measured flexor and extensor chamber pressures and the measured external
force input. Because this baseline requires measured chamber pressures as
inputs, it was evaluated only for displacement prediction.

\subsubsection*{2) Koopman/EDMDc lifted-regression model}

The Koopman/EDMDc baseline was implemented as a discrete-time lifted-regression
model. The physical state was
\[
\mathbf{x}_k =
\begin{bmatrix}
x_k & \dot{x}_k & P_{f,k} & P_{e,k}
\end{bmatrix}^{\top},
\]
and the external force was used as the input. 
The state and input were standardized using statistics fitted over the training
datasets. The lifting dictionary was applied to the standardized state
\(\tilde{\mathbf{x}}_k\). The default second-order dictionary is written
compactly as
\[
\psi(\tilde{\mathbf{x}}_k)=
\begin{bmatrix}
1 \\
\mathbf{x}_k \\
\phi_2(\mathbf{x}_k) \\
P_{f,k}-P_{e,k} \\
P_{f,k}+P_{e,k}
\end{bmatrix},
\]
where
\[
\begin{aligned}
\phi_2(\mathbf{x}_k)=
\big[
&x_k^2,\ \dot{x}_k^2,\ P_{f,k}^2,\ P_{e,k}^2,\\
&x_k\dot{x}_k,\ x_kP_{f,k},\ x_kP_{e,k},\\
&\dot{x}_kP_{f,k},\ \dot{x}_kP_{e,k},\ P_{f,k}P_{e,k}
\big]^\top .
\end{aligned}
\]
In normalized coordinates, the lifted regression model was trained by ridge
regression:
\[
\tilde{\mathbf{x}}_{k+1}
=
K
\begin{bmatrix}
\psi(\tilde{\mathbf{x}}_k)\\
\tilde{u}_k
\end{bmatrix},
\]
where \(K\) is the fitted linear map and \(\tilde{u}_k\) is the standardized
external force input. During rollout, the predicted physical
state was recursively relifted at each step before predicting the next state.
This relifted rollout was used to improve numerical stability compared with a
pure lifted-state rollout.

\subsubsection*{3) GRU recurrent model}

The GRU baseline was implemented as a recurrent multi-step predictor.
\revnew{The input sequence contained the measured state history, the external
force, and the two chamber air masses of the operating condition, which are
constant within a recording.} The
state vector was
\[
\mathbf{x}_k =
\begin{bmatrix}
x_k & \dot{x}_k & P_{f,k} & P_{e,k}
\end{bmatrix}^{\top}.
\]
Input and output variables were scaled to \([-1,1]\) using MinMax scaling fitted
over the training datasets. The network used two GRU layers with hidden size
96, followed by a fully connected prediction head. A 100-sample measured history
window was used for warm-up, after which predictions were generated recursively.
\revnew{The recursive training horizon was set to 500 samples.}
The training loss was applied to the scaled output variables using the same
relative state weighting as the Neural ODE training objective, with joint
displacement and velocity weighted more strongly than pressure:
\[
\begin{aligned}
\mathcal{L}_{\mathrm{GRU}}
=
\frac{1}{N}\sum_{k=1}^{N}
\Big[
&100(\hat{\tilde{x}}_k-\tilde{x}_k)^2
+100(\hat{\tilde{\dot{x}}}_k-\tilde{\dot{x}}_k)^2 \\
&+(\hat{\tilde{P}}_{f,k}-\tilde{P}_{f,k})^2
+(\hat{\tilde{P}}_{e,k}-\tilde{P}_{e,k})^2
\Big],
\end{aligned}
\]
where tildes denote MinMax-scaled variables. The model was trained using Adam 
with gradient clipping and early stopping.

\sisection{Baseline Comparison}\label{si:baseline_cmp}

\revnew{Table~\ref{tab:model_class_comparison} summarizes the roles of the model classes
compared in Section~\ref{subsec:forward_model}.}

\begin{table}[!htb]
\centering
\caption{Qualitative comparison of baseline model classes. ``Force'' and ``Stiffness'' indicate direct force representation and stiffness extraction capability.}
\label{tab:model_class_comparison}
\small
\setlength{\tabcolsep}{4pt}
\renewcommand{\arraystretch}{1.3}
\begin{tabular}{l c c c c}
\hline
Model type & Prediction & Force & Stiffness & Interpretability \\
\hline
Analytical & Weak & Strong & Strong & High \\
Koopman/EDMDc & Strong & Indirect & Indirect & Moderate \\
GRU & Strong & Weak & Indirect & Low \\
Neural ODE & Strong & Strong & Strong & High \\
\hline
\end{tabular}
\end{table}

\begin{figure}[!htb]
    \centering
    \includegraphics[width=\textwidth]{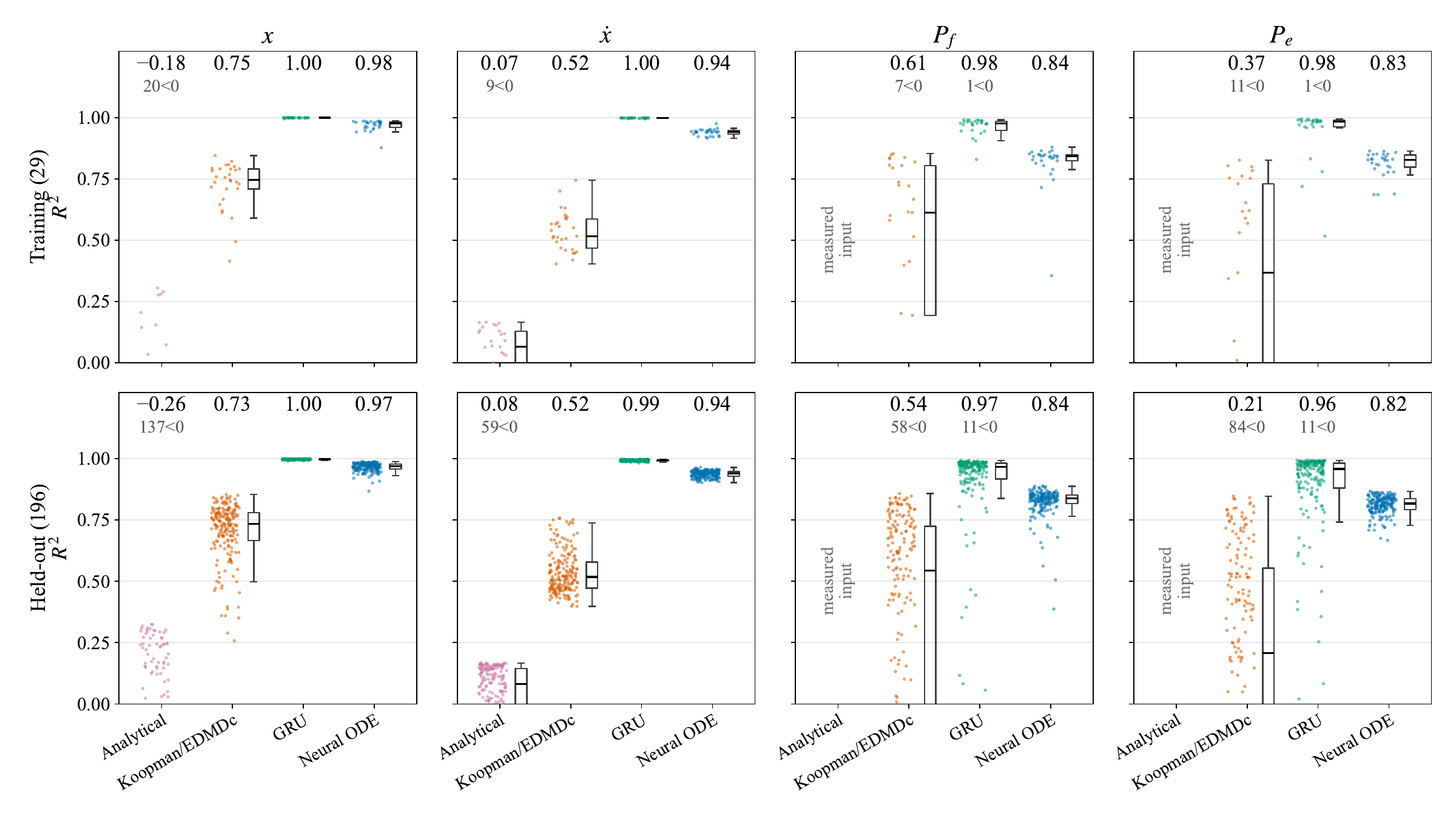}
    \caption{%
    Per-state \(R^2\) for all models on the 29 training (top) and 196 held-out
    (bottom) conditions. Boxes show median and interquartile range; dots are
    individual conditions; numbers above are medians over all conditions, with
    the count of negative values (not plotted) below; a box is omitted where
    the median is negative. The analytical (Kang-type) model uses measured
    chamber pressures as inputs and predicts \(x\) and \(\dot{x}\) only. The GRU
    is initialized from 100 measured samples; the other models start from a
    single measured state.}
    \label{fig:si_full_r2}
\end{figure}

\revnew{Fig.~\ref{fig:si_full_r2} and Table~\ref{tab:si_full_r2} report $R^2$
for each state and each model over the 29 training and 196 held-out
conditions. The median held-out pressure $R^2$ of the GRU (0.97 and 0.96 for
$P_f$ and $P_e$) is higher than that of the Neural ODE (0.84 and 0.82), but the
GRU gave a negative $R^2$ for at least one chamber pressure in 18 of the 196
held-out conditions (11 for each chamber, minimum $-7.4$), whereas the lowest
Neural ODE values were 0.39 and 0.67.
In absolute terms, the median held-out errors of the Neural ODE were 0.13~mm
in displacement and 4.2 and 4.3~kPa in the two pressures (GRU: 0.04~mm, 1.9
and 2.1~kPa); the largest pressure error was 8.5~kPa for the Neural ODE and
13.8~kPa for the GRU.}

\begin{table}[!htb]
\centering
\caption{Per-state $R^2$ and RMSE of the hybrid Neural ODE and the baselines over the 29 training and 196 held-out conditions. Each entry is the median over conditions with the worst condition in parentheses (minimum $R^2$, maximum RMSE); the interquartile range is given in brackets on the line below. The $R^2$ of $x$ is the metric of Fig.~\ref{fig:comparison}. ``Four-state'' is the $R^2$ averaged over $x$, $\dot{x}$, $P_f$, and $P_e$ for each condition; its median over the held-out conditions is the 0.89 reported in the main text (mean 0.88).}
\label{tab:si_full_r2}
\begin{threeparttable}
\small
\setlength{\tabcolsep}{4.5pt}
\begin{tabular}{llccccc}
\toprule
Model & Set & $x$ & $\dot{x}$ & $P_f$ & $P_e$ & Four-state \\
\midrule
\multicolumn{7}{l}{\textit{$R^2$: median (minimum); [interquartile range]}} \\
\midrule
Analytical & train & $-$0.181 ($-$5.82) & 0.065 ($-$0.11) & --- & --- & --- \\
 & & [$-$0.750--0.074] & [$-$0.007--0.129] &  &  &  \\
 & held-out & $-$0.256 ($-$6.25) & 0.082 ($-$0.13) & --- & --- & --- \\
 & & [$-$1.013--0.124] & [$-$0.022--0.144] &  &  &  \\
\midrule
Koopman/EDMDc & train & 0.745 (0.41) & 0.516 (0.40) & 0.613 ($-$10.61) & 0.367 ($-$36.34) & 0.523 ($-$8.63) \\
 & & [0.709--0.791] & [0.468--0.586] & [0.193--0.804] & [$-$0.335--0.730] & [$-$0.242--0.646] \\
 & held-out & 0.734 (0.26) & 0.518 (0.40) & 0.544 ($-$15.70) & 0.208 ($-$30.33) & 0.355 ($-$7.21) \\
 & & [0.666--0.780] & [0.472--0.579] & [$-$0.404--0.724] & [$-$1.457--0.553] & [$-$0.677--0.610] \\
\midrule
GRU & train & 0.999 (0.99) & 0.998 (0.99) & 0.977 ($-$0.74) & 0.983 ($-$1.14) & 0.990 (0.45) \\
 & & [0.999--0.999] & [0.998--0.998] & [0.948--0.985] & [0.962--0.988] & [0.975--0.991] \\
 & held-out & 0.997 (0.99) & 0.994 (0.98) & 0.967 ($-$5.08) & 0.958 ($-$7.39) & 0.975 ($-$2.62) \\
 & & [0.996--0.998] & [0.992--0.995] & [0.917--0.981] & [0.881--0.982] & [0.937--0.986] \\
\midrule
Neural ODE & train & 0.977 (0.88) & 0.942 (0.92) & 0.842 (0.36) & 0.828 (0.69) & 0.896 (0.75) \\
 & & [0.960--0.981] & [0.933--0.948] & [0.824--0.849] & [0.798--0.847] & [0.876--0.901] \\
 & held-out & 0.969 (0.87) & 0.940 (0.90) & 0.837 (0.39) & 0.817 (0.67) & 0.888 (0.77) \\
 & & [0.958--0.978] & [0.928--0.947] & [0.817--0.851] & [0.793--0.837] & [0.877--0.898] \\
\midrule
\multicolumn{7}{l}{\textit{RMSE in mm ($x$), mm/s ($\dot{x}$) and kPa ($P_f$, $P_e$): median (maximum); [interquartile range]}} \\
\midrule
Analytical & train & 0.818 (1.799) & 6.26 (10.54) & --- & --- &  \\
 & & [0.753--0.980] & [6.18--6.38] &  &  &  \\
 & held-out & 0.838 (1.835) & 6.24 (9.36) & --- & --- &  \\
 & & [0.757--1.030] & [6.01--6.35] &  &  &  \\
\midrule
Koopman/EDMDc & train & 0.366 (0.668) & 4.37 (5.77) & 6.67 (16.56) & 8.58 (25.56) &  \\
 & & [0.350--0.410] & [4.30--4.40] & [5.13--9.08] & [5.25--13.48] &  \\
 & held-out & 0.378 (0.598) & 4.37 (5.10) & 7.75 (20.56) & 10.15 (26.50) &  \\
 & & [0.355--0.424] & [4.31--4.41] & [5.90--9.61] & [6.87--13.97] &  \\
\midrule
GRU & train & 0.023 (0.053) & 0.27 (0.48) & 1.60 (5.49) & 1.39 (6.11) &  \\
 & & [0.019--0.024] & [0.25--0.30] & [1.16--2.14] & [1.21--1.73] &  \\
 & held-out & 0.038 (0.097) & 0.51 (1.13) & 1.93 (13.81) & 2.08 (13.46) &  \\
 & & [0.034--0.044] & [0.46--0.59] & [1.52--2.71] & [1.52--3.14] &  \\
\midrule
Neural ODE & train & 0.118 (0.242) & 1.52 (2.15) & 4.17 (8.29) & 4.22 (8.64) &  \\
 & & [0.097--0.162] & [1.46--1.64] & [3.42--5.26] & [3.22--5.42] &  \\
 & held-out & 0.127 (0.248) & 1.54 (2.35) & 4.17 (6.98) & 4.31 (8.54) &  \\
 & & [0.108--0.162] & [1.42--1.76] & [3.15--5.43] & [3.04--5.78] &  \\
\bottomrule
\end{tabular}
\begin{tablenotes}
\footnotesize
\item All models were trained on the same 29 conditions; the Koopman/EDMDc and GRU models were trained on all four states. The analytical model uses measured chamber pressures as inputs and predicts $x$ and $\dot{x}$ only. The GRU is initialized from 100 measured samples; the other models start from a single measured state.
\end{tablenotes}
\end{threeparttable}
\end{table}

\begin{figure}[!htb]
    \centering
    \begin{minipage}[t]{0.49\textwidth}
        \raggedright\textbf{a}\\[0.6ex]
        \centering
        \includegraphics[width=\linewidth]{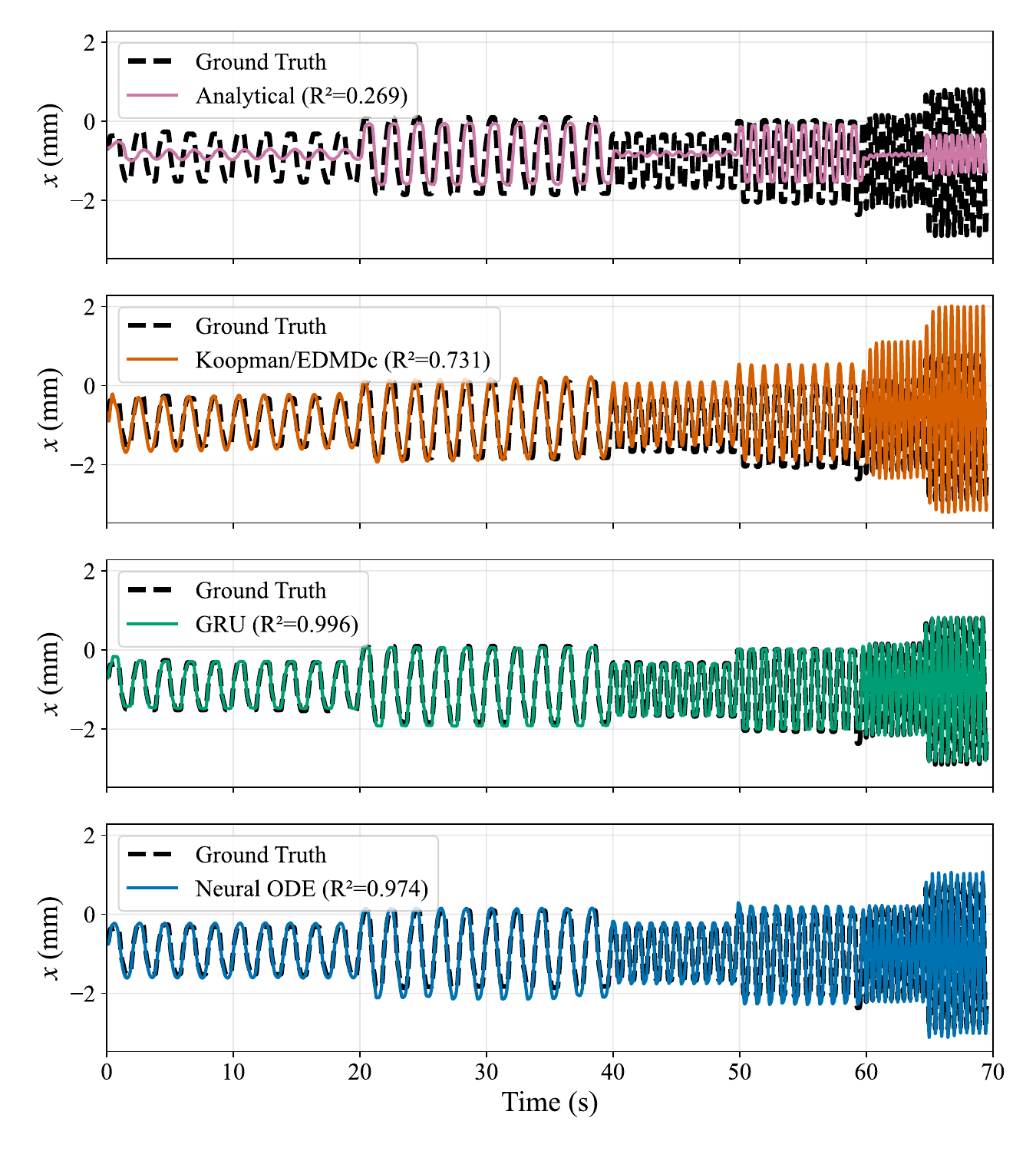}
    \end{minipage}%
    \hfill
    \begin{minipage}[t]{0.49\textwidth}
        \raggedright\textbf{b}\\[0.6ex]
        \centering
        \includegraphics[width=\linewidth]{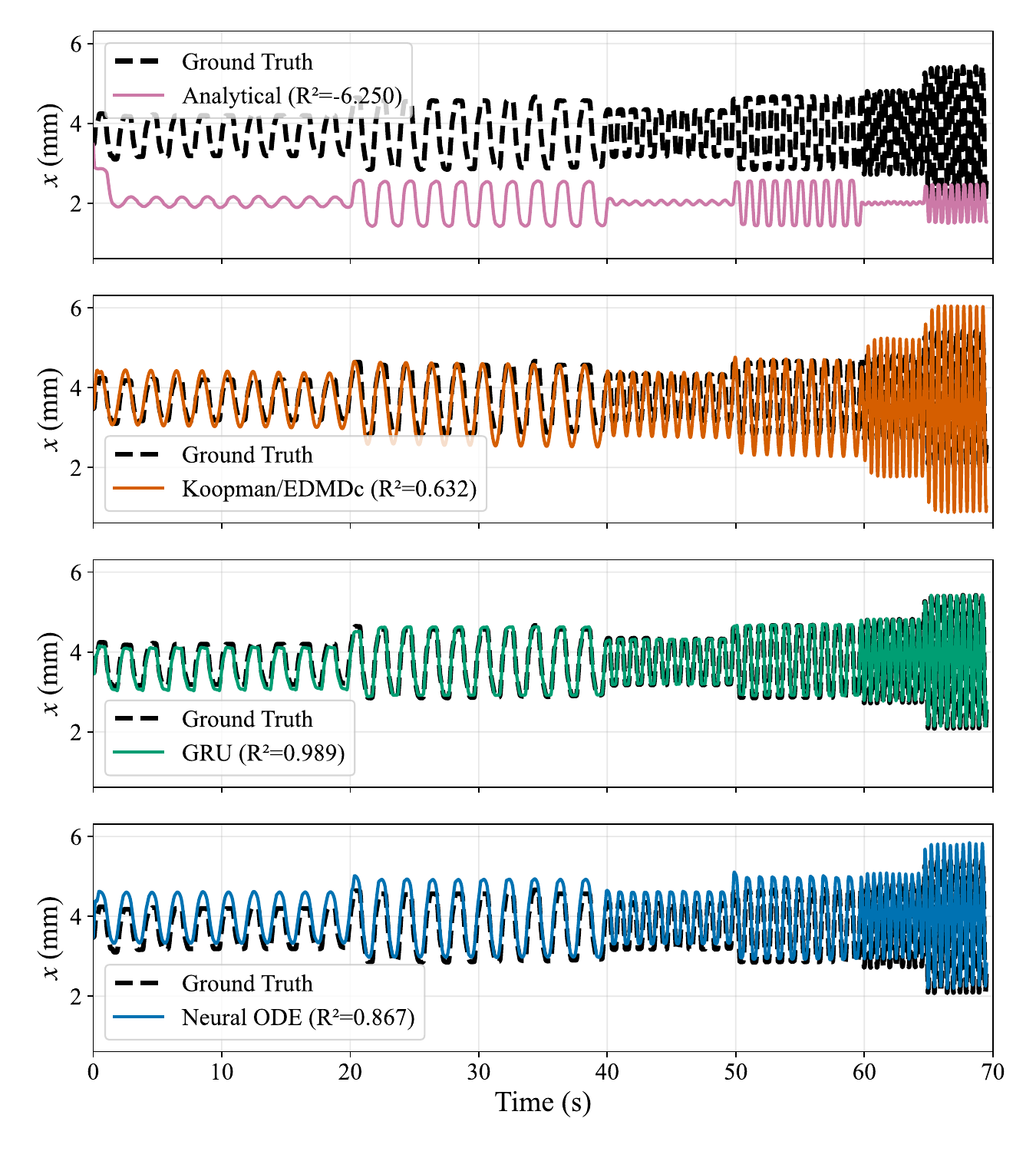}
    \end{minipage}
    \caption{%
    Displacement response comparison for two held-out conditions, not used to
    train any model. \textbf{a} The condition of
    Fig.~\ref{fig:response_example}, 308-446~kPa absolute (30-50~psi gauge).
    \textbf{b} 653-205~kPa absolute (80-15~psi gauge), the held-out condition
    with the lowest Neural ODE displacement \(R^2\) (0.867); the Neural ODE
    follows the motion with an offset, and the Kang-type baseline does not
    reach the operating point.
    Reported \(R^2\) values are displacement-only scores over
    the full trajectory. The Kang-type baseline uses measured chamber pressures,
    while the Koopman/EDMDc, GRU, and hybrid Neural ODE models use recursive rollout.
    The GRU is initialized from 100 measured samples; the other models start
    from a single measured state.}
    \label{fig:supp_displacement_comparison}
\end{figure}

\revnew{Fig.~\ref{fig:supp_displacement_comparison}{\textbf{a}} compares the displacement
response of the hybrid Neural ODE and the baseline models for the held-out
condition of Fig.~\ref{fig:response_example} (30-50~psi gauge).} The \(R^2\) values shown in
the legend are computed for the displacement trajectory over the full time
window.
\revnew{Fig.~\ref{fig:supp_displacement_comparison}{\textbf{b}} shows the same
comparison for the held-out condition with the lowest Neural ODE displacement
$R^2$.}

\sisection{Tracking of Sinusoidal Trajectories}\label{si:tracking_sine}

\begin{figure}[!htb]
    \centering
    \includegraphics[width=0.33\textwidth]{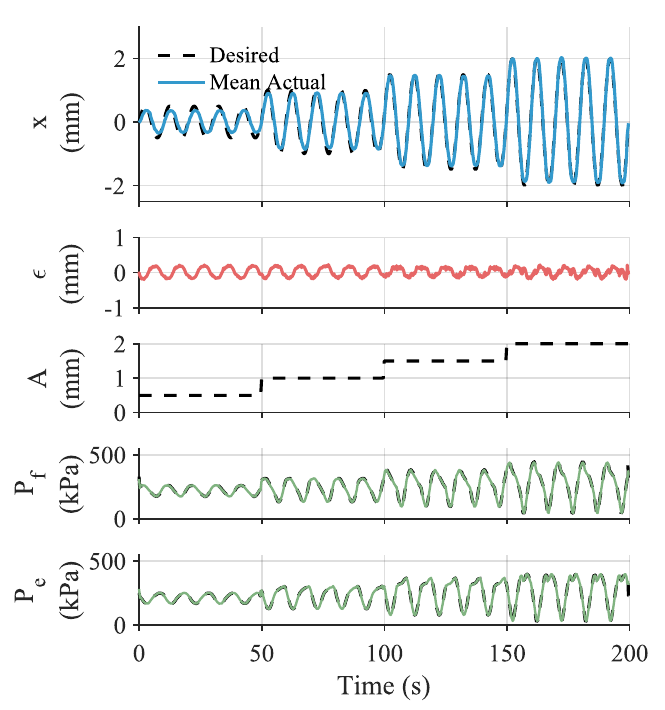}%
    \hfill
    \includegraphics[width=0.33\textwidth]{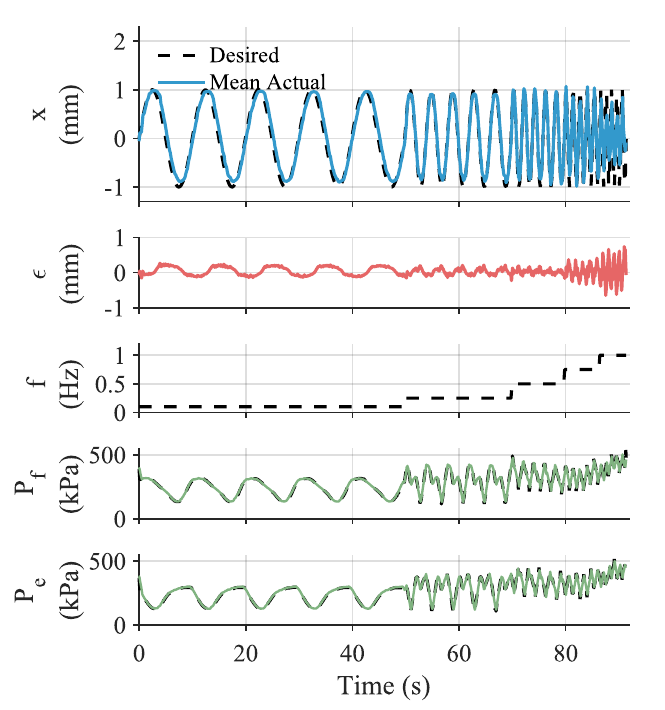}%
    \hfill
    \includegraphics[width=0.33\textwidth]{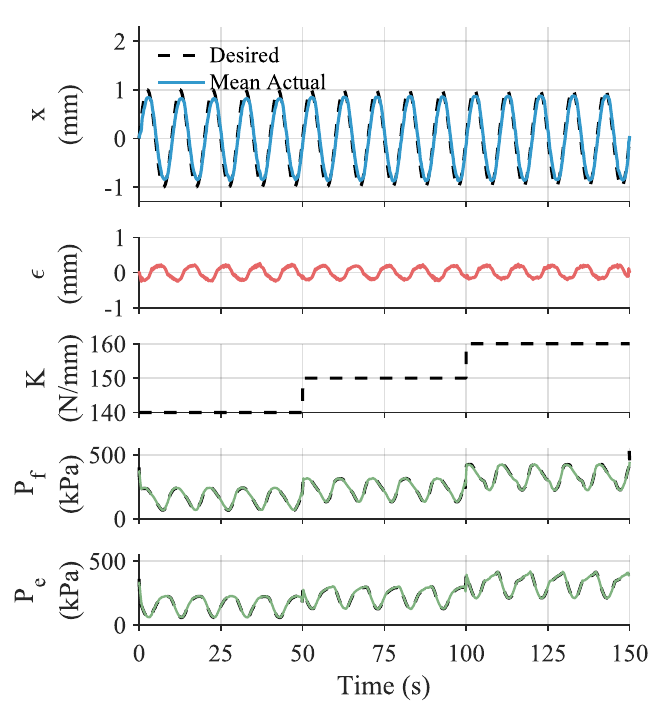}
    \put (-500,170){\textbf{a}}
    \put (-330,170){\textbf{b}}
    \put (-155,170){\textbf{c}}
    \caption{%
    Feedforward tracking validation for sinusoidal trajectories under
    {\textbf{a}} amplitude variation ($A=\{0.5,1.0,1.5,2.0\}$~mm),
    {\textbf{b}} frequency variation ($f=\{0.1,0.25,0.5,0.75,1.0\}$~Hz), and
    {\textbf{c}} stiffness variation ($K_d=\{140,150,160\}$~N/mm). Traces are means
    of six repetitions.}
    \label{fig:si_tracking_sine}
\end{figure}

\revnew{Fig.~\ref{fig:si_tracking_sine} shows the tracking experiments of
Section~\ref{subsec:results_inverse} for sinusoidal trajectories, under the same
amplitude, frequency, and stiffness conditions as the triangular trajectories
of Fig.~\ref{fig:tracking_performance}.}

\sisection{Simulated Tracking Error Above 1~Hz}\label{si:simfreq}

\revnew{Tracking was measured up to 1~Hz. To indicate how imperfect pressure
tracking affects the tracking error from 1.0 to 1.5~Hz, commands were
synthesized for a 1-mm sinusoid at $K_d=150$~N/mm and the model was driven with
them in two cases: delayed by 28~ms, and delayed by 28~ms with the air-mass
variation scaled by 0.81. These are the flexor pressure lag and amplitude ratio
measured in the 1-Hz tracking experiment, applied here to both chambers and at
every frequency. The commands are synthesized from the same model that is then
simulated, so exact commands give zero error by construction (below
$10^{-4}$~mm). Table~\ref{tab:si_simfreq} therefore shows the sensitivity of
the model to delay and attenuation and is not a prediction of the error on the
hardware: at 1~Hz it gives 0.20~mm where the experiment gave 0.37~mm.}

\begin{table}[!htb]
\centering
\caption{Simulated tracking RMSE (mm) for a 1-mm sinusoid at $K_d=150$~N/mm when the synthesized commands are delayed and attenuated, and the measured value at 1~Hz.}
\label{tab:si_simfreq}
\small
\begin{tabular}{lccccc}
\toprule
 & 1.0~Hz & 1.2~Hz & 1.3~Hz & 1.4~Hz & 1.5~Hz \\
\midrule
28-ms delay                 & 0.12 & 0.15 & 0.16 & 0.17 & 0.19 \\
28-ms delay, 0.81 amplitude & 0.20 & 0.22 & 0.23 & 0.24 & 0.25 \\
Measured                    & 0.37 & --- & --- & --- & --- \\
\bottomrule
\end{tabular}
\end{table}

\sisection{Tracking Error by Condition}\label{si:rmse}

\revnew{Table~\ref{tab:si_rmse} gives the tracking RMSE of every condition of
Section~\ref{subsec:results_inverse}, as the mean and standard deviation over
the six repetitions.}

\begin{table}[!htb]
\centering
\caption{Tracking RMSE (mm), mean $\pm$ SD over six repetitions. In each test the other two quantities are fixed at $A=1.0$~mm, $f=0.1$~Hz, and $K_d=150$~N/mm.}
\label{tab:si_rmse}
\small
\begin{tabular}{llcc}
\toprule
Varied & Value & Sinusoidal & Triangular \\
\midrule
Amplitude (mm) & 0.5 & $0.136\pm0.008$ & $0.104\pm0.003$ \\
 & 1.0 & $0.131\pm0.004$ & $0.111\pm0.001$ \\
 & 1.5 & $0.110\pm0.002$ & $0.121\pm0.003$ \\
 & 2.0 & $0.108\pm0.001$ & $0.143\pm0.003$ \\
\midrule
Frequency (Hz) & 0.1 & $0.135\pm0.004$ & $0.147\pm0.011$ \\
 & 0.25 & $0.092\pm0.002$ & $0.077\pm0.001$ \\
 & 0.5 & $0.069\pm0.006$ & $0.118\pm0.002$ \\
 & 0.75 & $0.189\pm0.002$ & $0.272\pm0.017$ \\
 & 1.0 & $0.374\pm0.003$ & $0.324\pm0.025$ \\
\midrule
Stiffness (N/mm) & 140 & $0.160\pm0.010$ & $0.139\pm0.004$ \\
 & 150 & $0.147\pm0.003$ & $0.134\pm0.002$ \\
 & 160 & $0.143\pm0.002$ & $0.136\pm0.003$ \\
\bottomrule
\end{tabular}
\end{table}

\sisection{Commanded and Identified Stiffness}\label{si:command}

\begin{figure}[!htb]
    \centering
    \includegraphics[width=\textwidth]{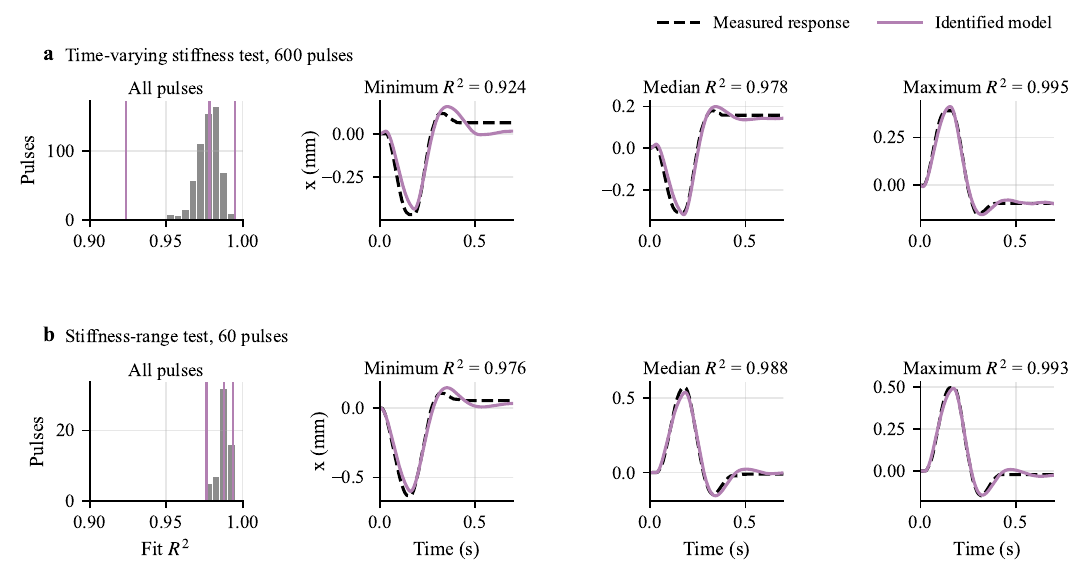}
    \caption{%
    Quality of the stiffness identification. \textbf{a} Time-varying stiffness
    test (Fig.~\ref{fig:stiffness_verification}\textbf{a}); \textbf{b}
    stiffness-range test (Fig.~\ref{fig:stiffness_verification}\textbf{b}).
    In each row, the histogram shows the fit \(R^2\) over all pulses, with
    vertical lines at the minimum, median, and maximum, and the three traces
    show the measured response and the identified model for the pulses with the
    minimum, median, and maximum \(R^2\).}
    \label{fig:si_sysid}
\end{figure}

\revnew{The stiffness values of Section~\ref{subsec:stiffness} are obtained by fitting a
spring--mass--damper model to the response to each pulse.
Fig.~\ref{fig:si_sysid} shows the coefficient of determination of that fit for
every pulse, and the fits with the lowest, median, and highest values. Over
the 600 pulses of the time-varying stiffness test the median $R^2$ was 0.978
(minimum 0.924); over the 60 pulses of the stiffness-range test it was 0.988
(minimum 0.976).}

\begin{figure}[!htb]
    \centering
    \includegraphics[width=\textwidth]{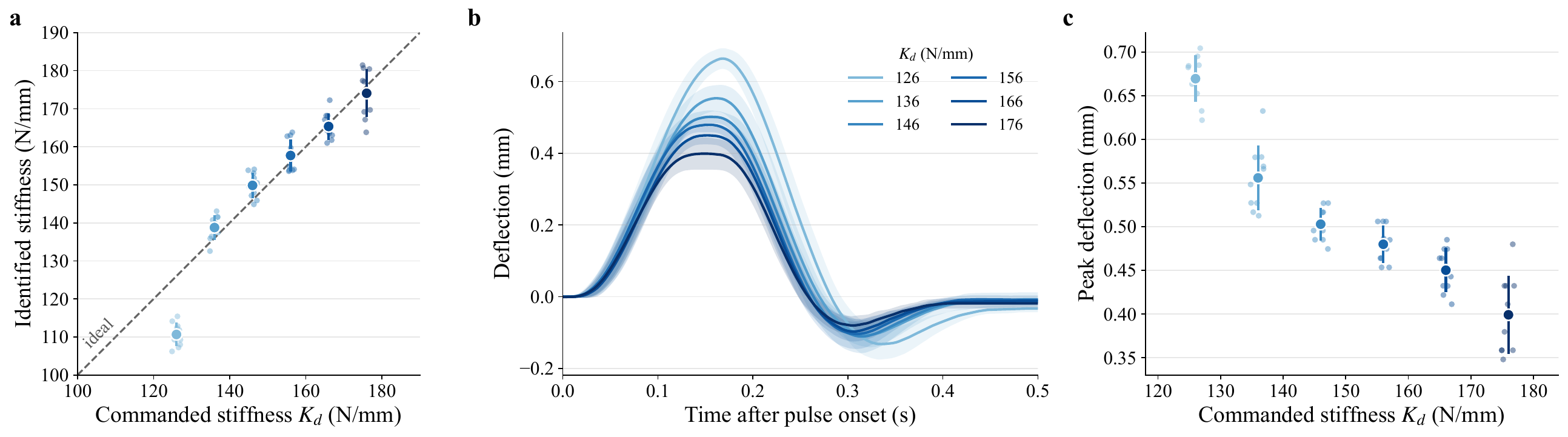}
    \caption{%
    Stiffness-range verification (six commanded levels, ten pulses each, joint
    at equilibrium, valves closed). \textbf{a} Identified versus commanded
    stiffness; small dots are individual pulses, large dots are mean \(\pm\)
    SD, and the dashed line is identity. \textbf{b} Mean \(\pm\) SD deflection
    after onset of the same torque pulse. \textbf{c} Peak deflection versus
    commanded stiffness.}
    \label{fig:si_command}
\end{figure}

\revnew{Fig.~\ref{fig:si_command} summarizes all 60 pulses of the
stiffness-range experiment in Section~\ref{subsec:stiffness} (six commanded
levels, ten pulses each). From 136 to 176~N/mm, the mean identified stiffness
of each level was within 3.9~N/mm of the commanded value (2.2~N/mm on
average); at 126~N/mm it was 110.7~N/mm. Under the same torque pulse, the mean
peak deflection decreased monotonically from 0.67 to 0.40~mm across the
commanded range.}

\sisection{Comparison with the Equilibrium-Point Model}\label{si:ep}

\revnew{Table~\ref{tab:stiffness_value_and_percentage} gives the identified stiffness
behind Fig.~\ref{fig:NN_EP_comparison} and the comparison of
Section~\ref{subsec:EP_comparison}.}

\begin{table}[!htb]
\centering
\caption{Identified stiffness (N/mm) for the Neural ODE (NN) and equilibrium-point (EP) models at zero and maximum velocity.}
\label{tab:stiffness_value_and_percentage}
\begin{threeparttable}
\setlength{\tabcolsep}{4pt}
\begin{tabular}{ccccccc}
\toprule
Model & Part & \shortstack{Nominal\\(N/mm)} & $\dot{x}=0$ & $\dot{x}=\dot{x}_{\max}$ & $\Delta(\%)$ & $p$-value \\
\midrule
\multirow{3}{*}{NN}
 & K$_{\mathrm{low}}$  & 140 & 139.54 & 140.07 & 1.06 & 0.6036 \\
 & K$_{\mathrm{mid}}$  & 150 & 146.96 & 145.83 & 2.26 & 0.1075 \\
 & K$_{\mathrm{high}}$ & 160 & 152.39 & 153.36 & 1.94 & 0.2585 \\
\midrule
\multirow{3}{*}{EP}
 & K$_{\mathrm{low}}$  & 140 & 140.66 & 136.64 & 8.04 & $3.9\times10^{-12}$ \\
 & K$_{\mathrm{mid}}$  & 150 & 146.98 & 143.09 & 7.78 & $1.9\times10^{-13}$ \\
 & K$_{\mathrm{high}}$ & 160 & 151.39 & 147.46 & 7.86 & $9.3\times10^{-14}$ \\
\bottomrule
\end{tabular}
\begin{tablenotes}
\footnotesize
\item Stiffness values are identified at zero velocity ($\dot{x}=0$) and maximum velocity ($\dot{x}=\dot{x}_{\max}$).
$\Delta$ denotes the absolute change in stiffness between
$\dot{x}=0$ and $\dot{x}=\dot{x}_{\max}$, normalized by the
theoretical stiffness range (126--176~N/mm).
$p$-values are obtained using paired $t$-tests comparing the stiffness
identified at $\dot{x}=0$ and $\dot{x}=\dot{x}_{\max}$.
\end{tablenotes}
\end{threeparttable}
\end{table}

\sisection{Feedforward Load Holding}\label{si:payload}

\revnew{The pressure plans of the load-holding task were computed for commanded
payload peaks of 100, 150, and 165~N; the peaks measured from the motor current
were 107, 158, and 173~N. The uncompensated plan was run twice, with the
pressure controller active and with the valves closed.
Table~\ref{tab:si_payload} gives the displacement in the load direction at peak
payload and after the payload was removed, and Table~\ref{tab:si_payload_pred}
compares the uncompensated displacement with the prediction of the model.}

\begin{table}[!htb]
\centering
\caption{Feedforward load holding at a commanded stiffness of 150~N/mm: displacement in the load direction (mm), mean $\pm$ SD of 10 trials at peak payload, and range of the three means after unloading.}
\label{tab:si_payload}
\begin{threeparttable}
\small
\begin{tabular}{lcccc}
\toprule
 & \multicolumn{3}{c}{At peak payload (commanded)} & After \\
\cmidrule(lr){2-4}
Plan & 100~N & 150~N & 165~N & unloading \\
\midrule
No compensation, valves closed       & $0.57\pm0.03$ & $0.91\pm0.06$ & $0.96\pm0.06$ & 0.26--0.34 \\
No compensation, pressure controller & $0.64\pm0.08$ & $1.00\pm0.07$ & $1.07\pm0.08$ & 0.32--0.41 \\
EP                                   & $0.12\pm0.05$ & $0.18\pm0.07$ & $0.19\pm0.07$ & 0.02--0.04 \\
Neural ODE                           & $0.04\pm0.06$ & $0.07\pm0.06$ & $0.05\pm0.06$ & 0.03--0.04 \\
\bottomrule
\end{tabular}
\end{threeparttable}
\end{table}

\begin{table}[!htb]
\centering
\caption{Model prediction of the uncompensated displacement in the load-holding task. Measured values are means of 10 trials.}
\label{tab:si_payload_pred}
\begin{threeparttable}
\small
\begin{tabular}{lccccc}
\toprule
 & Commanded & Measured peak & Measured & Model & \\
Run & payload (N) & payload (N) & displacement (mm) & prediction$^{a}$ (mm) & Difference \\
\midrule
Pressure controller & 100 & 107.9 & 0.64 & 0.73 & $+14\%$ \\
                    & 150 & 158.7 & 1.00 & 1.07 & $+7\%$ \\
                    & 165 & 172.7 & 1.07 & 1.18 & $+10\%$ \\
\midrule
Valves closed       & 100 & 106.7 & 0.57 & 0.65 & $+13\%$ \\
                    & 150 & 159.7 & 0.91 & 0.96 & $+6\%$ \\
                    & 165 & 173.9 & 0.96 & 1.05 & $+9\%$ \\
\bottomrule
\end{tabular}
\begin{tablenotes}
\footnotesize
\item[$^{a}$] From the static force balance of the model at the measured peak payload of each run, with constant pressures (pressure controller) or constant air masses (valves closed). Nothing was fitted to these recordings. Differences are computed from unrounded values.
\end{tablenotes}
\end{threeparttable}
\end{table}

\end{document}